\documentclass[10pt,twocolumn,letterpaper]{article}

\usepackage[pagenumbers]{cvpr} %

\usepackage[utf8]{inputenc} %
\usepackage[T1]{fontenc}    %
\usepackage[dvipsnames]{xcolor}

\usepackage{algorithm}
\usepackage{amsfonts}       %
\usepackage{amsmath}
\usepackage{amssymb}
\usepackage{amsthm}
\usepackage{array}
\usepackage{bm}
\usepackage{booktabs}       %
\usepackage{caption}
\usepackage{colortbl}
\usepackage{empheq}
\usepackage{enumitem}
\usepackage{etoolbox}
\usepackage{float}
\usepackage{makecell}
\usepackage{mathrsfs}
\usepackage{mdframed}
\usepackage{microtype}      %
\usepackage{multirow}
\usepackage{multicol}
\usepackage{nccmath}
\usepackage{nicefrac}       %
\usepackage[noend]{algorithmic}
\usepackage{pifont}
\usepackage{ragged2e}
\usepackage{rotating}
\usepackage{subcaption}
\usepackage{tabularx}
\usepackage{tikz}
\usetikzlibrary{tikzmark}
\usepackage{times}
\usepackage{url}            %
\usepackage{wrapfig}
\usepackage{color, colortbl}
\usepackage{xspace}
\usepackage{thm-restate}
\usepackage{xfrac}

\newcounter{rowcntr}[table]
\renewcommand{\therowcntr}{\arabic{chapter}.\the\numexpr\arabic{table}+1.\arabic{rowcntr}}

\AtBeginEnvironment{tabular}{\setcounter{rowcntr}{0}}
\newcolumntype{H}{>{\setbox0=\hbox\bgroup}c<{\egroup}@{}}

\newcommand{\PreserveBackslash}[1]{\let\temp=\\#1\let\\=\temp}
\newcolumntype{C}[1]{>{\centering\arraybackslash}m{#1}}
\newcolumntype{R}[1]{>{\raggedleft\arraybackslash}m{#1}}
\newcolumntype{L}[1]{>{\raggedright\arraybackslash}m{#1}}

\newcommand{\comment}[1]{}

\def\xx{{\boldsymbol x}}

\def\aa{{\boldsymbol a}}

\def\VV{{\boldsymbol V}}
\def\UU{{\boldsymbol U}}

\definecolor{colorYes}{RGB}{51,160,44}
\definecolor{colorNo}{RGB}{228,26,28} %

\newcommand{\cmark}{\textcolor{colorYes}{\ding{51}}}%
\newcommand{\xmark}{\textcolor{colorNo}{\ding{55}}}%

\newcommand{\imagenet}{ImageNet}
\newcommand{\flickr}{Flickr30k}

\def\ourmethod{{MobileCLIP}}
\def\ourdataset{{DataCompDR}}
\def\ourdatasetOneB{{DataCompDR-1B}}
\def\ourdatasetTM{{DataCompDR-12M}}
\def\datacomp{{DataComp}}
\def\datacompOneB{{DataComp-1B}}
\def\datacompTM{{DataComp-12M}}
\def\ourimage{{MCi}}
\def\ourimagezero{{MCi0}}
\def\ourimageone{{MCi1}}
\def\ourimagetwo{{MCi2}}
\def\ourtextblock{{Text-RepMixer}}
\def\ourtext{{MCt}}
\def\ourtextzero{{MCt}}

\def\ourmodel{{MobileCLIP}}
\def\ourmodelB{{MobileCLIP-B}}
\def\ourmodelSTwo{{MobileCLIP-S2}} %
\def\ourmodelSOne{{MobileCLIP-S1}} %
\def\ourmodelSZero{{MobileCLIP-S0}} %

\newcommand{\LTotal}{{\mathcal{L}_{\text{Total}}}}
\newcommand{\LKD}{{\mathcal{L}_{\text{Distill}}}}
\newcommand{\LKDIT}{{\mathcal{L}_{\text{Distill}}^{\text{I2T}}}}
\newcommand{\LKDTI}{{\mathcal{L}_{\text{Distill}}^{\text{T2I}}}}

\newcommand{\LClip}{{\mathcal{L}_{\text{CLIP}}}}
\newcommand{\Similarity}{{\mathcal{S}}}

\newcommand{\FeatStudentImg}{{\varPhi_{\text{img}}}}
\newcommand{\FeatStudentTxt}{{\varPhi_{\text{txt}}}}

\newcommand{\FeatTeacherImgk}{{\Psi_{\text{img}}^{(k)}}}
\newcommand{\FeatTeacherTxtk}{{\Psi_{\text{txt}}^{(k)}}}

\newcommand{\imagei}{{\xx}_{\text{img}}^{(i)}}
\newcommand{\gttexti}{{\xx}_{\text{txt}}^{(i)}}
\newcommand{\paramsij}{{\aa}^{(i,j)}}
\newcommand{\augimageij}{{\hat{\xx}}_{\text{img}}^{(i,j)}}
\newcommand{\syntextis}{{\xx}_{\text{syn}}^{(i,s)}}

\newcommand{\FeatTeacherImgijk}{{\psi_{\text{img}}^{(i,j,k)}}}
\newcommand{\FeatTeacherTxtik}{{\psi_{\text{txt}}^{(i,k)}}}
\newcommand{\FeatTeacherSynisk}{{\psi_{\text{syn}}^{(i,s,k)}}}

\def\imagenetval{{IN-val}}
\def\flickrval{{Flickr30k}}

\definecolor{cvprblue}{rgb}{0.21,0.49,0.74}
\usepackage[pagebackref,breaklinks,colorlinks,citecolor=cvprblue]{hyperref}

\usepackage[capitalize]{cleveref}
\usepackage{pifont}
\crefname{section}{Sec.}{Secs.}
\Crefname{section}{Section}{Sections}
\Crefname{table}{Table}{Tables}
\crefname{table}{Tab.}{Tabs.}

\usepackage[textwidth=2.5cm]{todonotes}

\def\paperID{6874} %
\def\confName{CVPR}
\def\confYear{2024}

\title{\ourmethod{}: Fast Image-Text Models through Multi-Modal Reinforced Training
}

\author{Pavan Kumar Anasosalu Vasu\thanks{Equal contribution.}%
\and
Hadi Pouransari$^*$%
\and
Fartash Faghri$^*$%
\and
Raviteja Vemulapalli%
\and
Oncel Tuzel\\
Apple\\
{\tt\small \{panasosaluvasu,mpouransari,fartash,r\_vemulapalli,otuzel\}@apple.com}
}

\begin{document}
\maketitle
\begin{abstract}

Contrastive pretraining of image-text foundation models, such as CLIP, 
    demonstrated excellent zero-shot performance and improved robustness on 
    a wide range of downstream tasks.
    However, these models utilize large
    transformer-based encoders with 
    significant memory and latency overhead which pose challenges for deployment 
    on mobile devices.
    In this work, we introduce \ourmodel{} -- a new family 
    of efficient image-text models optimized for runtime performance along with 
    a novel and efficient training approach, namely 
    \textit{multi-modal reinforced training}.  The proposed training approach leverages 
    knowledge transfer from an image captioning model and an ensemble of strong 
    CLIP encoders to improve the accuracy of efficient models. Our approach avoids 
    train-time compute overhead by storing the additional knowledge in 
    a reinforced dataset.  \ourmodel{} sets a new state-of-the-art 
    latency-accuracy tradeoff for zero-shot classification and retrieval tasks 
    on several datasets. Our \ourmodelSTwo{} variant is 2.3$\times$ 
    faster while more accurate  
    compared to previous best CLIP model based on ViT-B/16. We further demonstrate 
    the
    effectiveness of our multi-modal reinforced training 
    by training a CLIP model based on ViT-B/16 image backbone and 
    achieving +2.9\% average performance improvement on 38 evaluation benchmarks compared to the previous 
    best. Moreover, we show that the proposed approach achieves 
    10$\times$-1000$\times$ improved learning efficiency when compared with 
    non-reinforced CLIP training. Code and models are available at 
    \url{https://github.com/apple/ml-mobileclip}

\end{abstract}
    
 \begin{figure}
    \centering
    \includegraphics[width=0.9\linewidth]{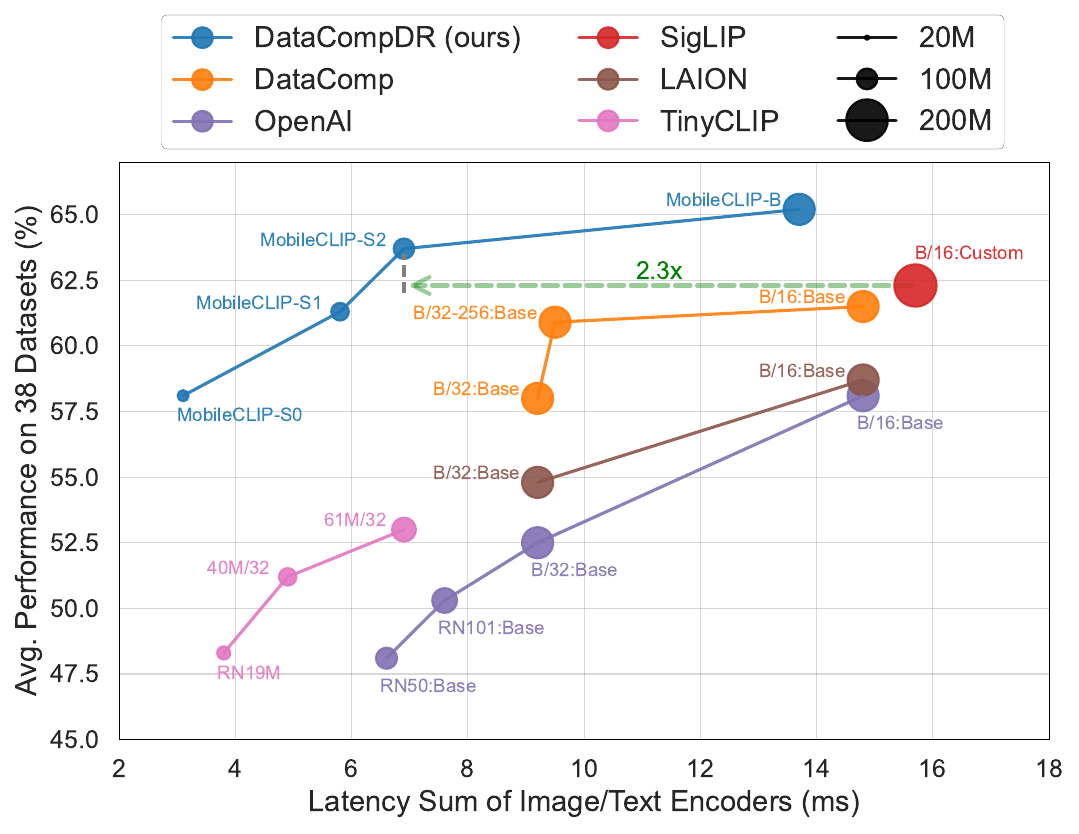}
    \vspace{-7pt}
    \caption{\textbf{\ourmodel{} models are fast and accurate.}
    Comparison of publicly available CLIP models with 
    \ourmethod{} trained on our \ourdataset{} dataset. Latency is measured 
    on iPhone12 Pro Max. %
    }
    \label{fig:accuracy_vs_latency_tradeoff}
\end{figure}

\begin{figure}
\vspace{-10pt}
    \centering
    \includegraphics[width=0.8\linewidth]{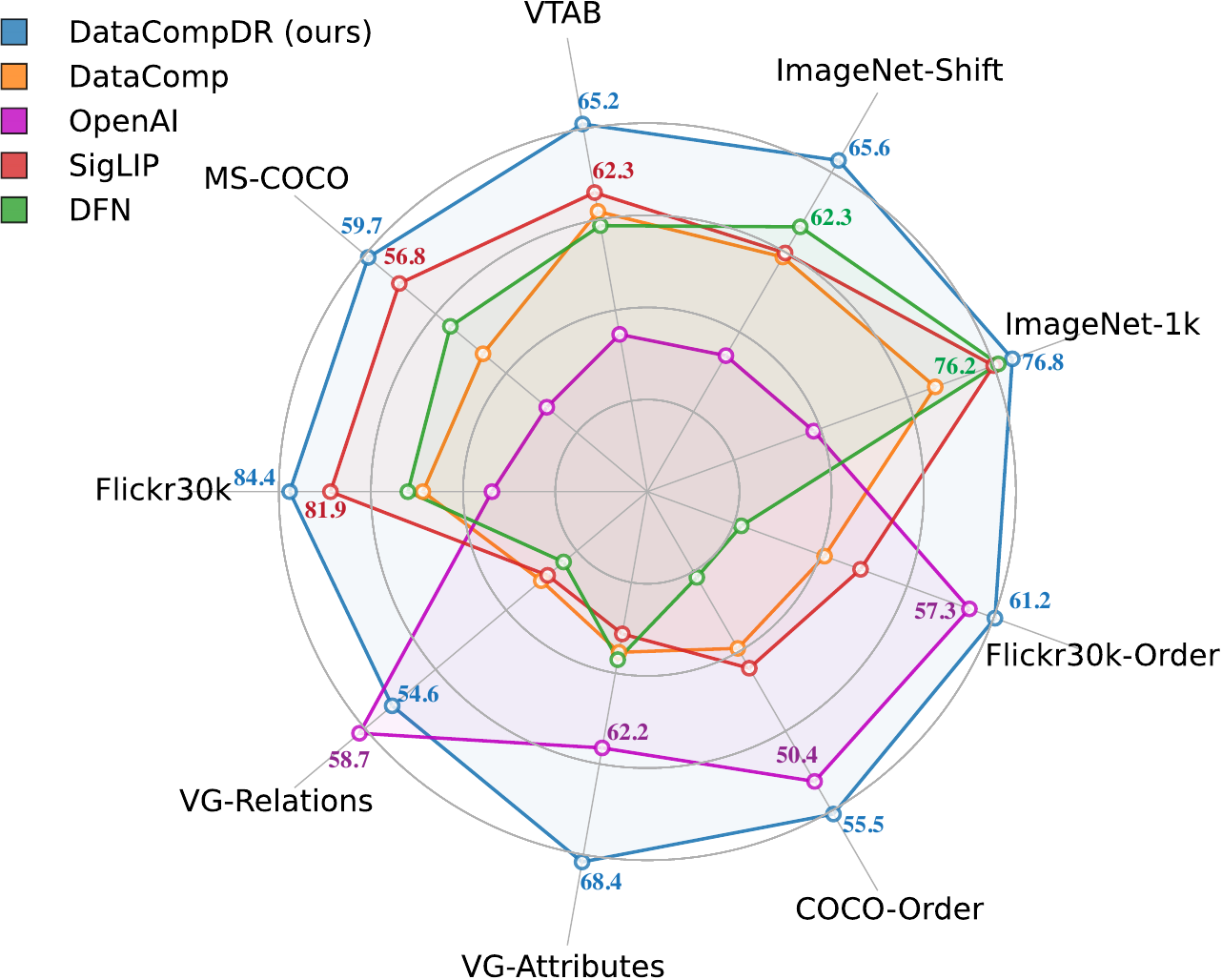}
    \vspace{-5pt}
    \caption{\textbf{\ourdataset{} dataset improves all metrics.}
    Zero-shot performance of CLIP models with ViT-B/16 image encoder.}
    \label{fig:accuracy_radar}
\vspace{-10pt}
\end{figure}

\section{Introduction}

Large image-text foundation models, such as CLIP~\citep{CLIP}, have demonstrated excellent zero-shot performance and improved robustness~\citep{Robustness} across a wide range of downstream tasks~\citep{ALIGN}. However, deploying these models on mobile devices is challenging due to their large size and high latency.

Our goal is to design a new family of aligned image-text encoders suitable for mobile devices. There are two main challenges towards realizing this goal. First, there is a tradeoff between runtime performance (e.g., latency) and the accuracy of different architectures, therefore we should be able to quickly and thoroughly analyze different architectural designs. Large-scale training of CLIP models is computationally expensive, hindering rapid development and exploration of efficient architecture design. On the other hand, standard multi-modal contrastive learning~\citep{CLIP} at small-scale results in poor accuracies, which do not provide a useful signal to guide architecture design choices. Second, reduced capacity of smaller architectures leads to subpar accuracy that can be improved with a better training method.

To overcome these challenges, we develop a novel training approach based on the dataset 
reinforcement method~\citep{DR}: i) reinforce a dataset once with additional 
information, and ii) use the reinforced dataset several times for 
experimentation. For a given compute budget, training with the reinforced 
dataset results in improved accuracy compared to the original dataset. We 
propose a multi-modal variant of dataset reinforcement for training efficient 
CLIP models. Specifically, we reinforce the image-text \datacomp{}~\citep{DataComp} dataset by adding synthetic captions and embeddings from a strong 
ensemble of pretrained CLIP models (\cref{fig:dr}), obtaining \ourdataset{}. We 
introduce two variants of our reinforced dataset, \ourdatasetTM{} suited for 
rapid iteration on efficient model design and 
\ourdatasetOneB{} for best large-scale 
training performance.

Training with \ourdataset{} shows significant learning efficiency improvement compared to 
the standard CLIP training. For example, with a single node of 8$\times$A100 GPUs, 
we achieve 61.7\% zero-shot classification on ImageNet-val~\citep{IN-1k} 
in approximately one day when training a ViT-B/16~\citep{ViT} based CLIP from 
scratch on \ourdatasetTM{}. Training with \ourdatasetOneB{} sets new 
state-of-the-art performance on several metrics (\cref{fig:accuracy_radar}) 
while still using a fraction of the training compute budget compared to previous works.

Utilizing \ourdataset{}, we explored the design space and obtained a new family of 
mobile-friendly aligned image-text encoders called \ourmodel{} with a better 
latency-accuracy tradeoff compared to the previous works 
(\cref{fig:accuracy_vs_latency_tradeoff}). We exploit several architectural 
design techniques to obtain efficient image and text encoders, including 
structural 
reparametrization~\cite{ding2019acnet,guo2020expandnets,ding2021diverse,repvgg,MobileOne} 
and convolutional token mixing~\cite{vasu2023fastvit}. \ourmodel{} includes S0, 
S1, S2, and B variants covering various sizes and latencies for different 
mobile applications. Our fastest variant, \ourmodelSZero{}, is approximately 
5$\times$ faster and 3$\times$ smaller than the standard OpenAI ViT-B/16 CLIP 
model~\citep{CLIP}, but has the same average accuracy. Our contributions are as follows:
\begin{itemize}
\item We design a new family of mobile-friendly CLIP models, \textit{\ourmodel{}}. Variants of \ourmodel{} use hybrid CNN-transformer architectures with structural reparametrization in image and text encoders to reduce the size and latency.

\item We introduce multi-modal reinforced training, a novel training strategy that incorporates knowledge transfer from a pre-trained image captioning model and an ensemble of strong CLIP models to improve learning efficiency.

\item We introduce two variants of our reinforced datasets: \ourdatasetTM{} and \ourdatasetOneB{}. Using \ourdataset{}, we demonstrate 10x-1000x learning efficiency in comparison to DataComp.
        
\item \ourmodel{} family obtains state-of-the-art latency-accuracy tradeoff on zero-shot tasks, including marking a new best ViT-B/16 based CLIP model.

\end{itemize}
\label{sec:intro}

\section{Related Work}
\paragraph{Efficient learning for CLIP.}
One can improve learning efficiency through utilizing an enhanced training objective. Examples include image masking~\citep{RILS,EVA,EVA-CLIP,FLIP}, unimodal self-supervision~\citep{SLIP,DeCLIP}, fine-grained image-text alignment~\citep{FILIP}, contrastive learning in image-text-label space~\citep{yang2022unified}, and pairwise Sigmoid loss~\citep{SigLIP}. CLIPA~\citep{CLIPA} proposed training at multi-resolutions for cost-effective training. These methods are complementary to our proposed method.

CLIP training dataset is often comprising 
noisy image-text pairs obtained at web-scale. Since the original CLIP 
model~\citep{CLIP}, several works have 
demonstrated improved results on large-scale and filtered datasets~\citep{Laion400,Laion,DataComp,SigLIP,DFN}. Complementary to data collection and filtering, recent works show that using visually enriched synthetic captions generated from a pretrained captioning model, along with real captions, can improve the quality of CLIP models~\citep{ALIP,SynCapt,VeCLIP}. 
Our proposed reinforced multi-modal dataset also benefits from synthetically 
generated captions, which we show are crucial for improved learning efficiency.

Previous works like DIME-FM~\citep{DIME-FM}, extends unimodal distillation~\citep{hinton2015distilling} with a focus on zero-shot classification.
TinyCLIP~\citep{tinyclip} trains compact CLIP models via cross-modal affinity mimicking and weight inheritance. 
Multi-modal distillation is also explored in setups where the student is a fused vision-language model for specific tasks~\citep{CLIP-TD,DLIP,MAD}. Our proposed multi-modal reinforced training also includes cross-modal affinity mimicking~\citep{tinyclip}. Further, we extend unimodal model ensembling~\citep{DeepEnsemble,XCL} to multimodal setup, and store targets obtained from an ensemble of CLIP models.

Offline knowledge distillation methods~\citep{shen2022fast,yun2021re,DR} have been proposed recently to mitigate the training-time overhead cost due to running large teacher models. We extend the \emph{dataset reinforcement} strategy~\citep{DR} to the multi-modal setup of CLIP. Our proposed reinforced multi-modal datasets result in significant accuracy improvement without adding a training-time computational overhead.

\vspace{-10pt}
\paragraph{Efficient architectures for CLIP.}
Recently there have been a wide range of architectures that have shown great promise for accomplishing vision tasks on resource constraint devices. These architectures can be broadly classified into purely convolutional~\citep{ResNet, ConvNext, RegNet, repvgg, MobileNet_v1, MobileNet_v2, Mobilenet_v3, MobileOne}, transformer based~\citep{ViT, Deit, Swin} and convolution-transformer hybrids like~\citep{efficientformer, efficientformerv2, vasu2023fastvit, FasterVit, swiftformer, mobilevig}. Similarly there are transformer based~\citep{transformers} and convolution-transformer hybrids like~\citep{lightconv, conformer} for text encoding. 
There have been works like~\citep{tinyclip}, that prune ViT architectures to obtain smaller and faster CLIP models or works like~\citep{pumer} that reduce image-text tokens for faster inference of vision-language models. These models can still be quite large and inefficient to be deployed on a mobile device. In our work, we introduce an improved convolution-transformer hybrid architecture for both vision and text modalities, that improve over recent state-of-the-art like~\citep{efficientformerv2, swiftformer, mobilevig, FasterVit}. The optimizations introduced in ~\citep{tinyclip, pumer} can be used to further improve efficiency of our models.

\section{Multi-Modal Reinforced Training}
\begin{figure*}
\centering
\begin{minipage}{.68\textwidth}
    \centering
    \includegraphics[width=1.0\linewidth,clip]{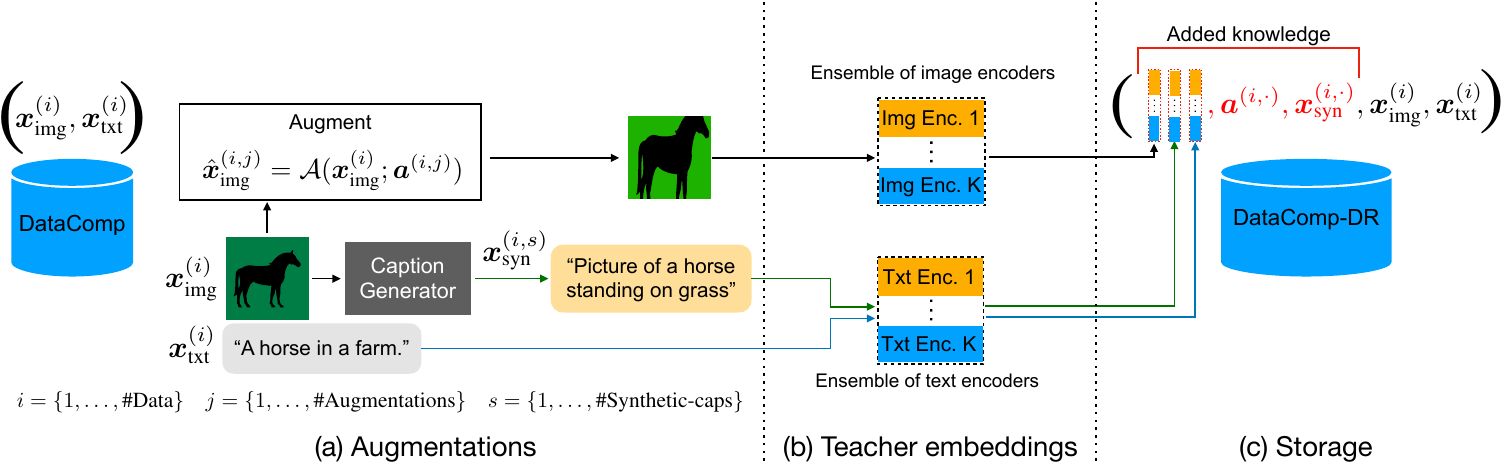}
    \captionof{figure}{Illustration of multi-modal dataset reinforcement with one image augmentation and one synthetic caption. In practice, we use multiple image augmentations and synthetic captions. 
    }
    \label{fig:dr}
\end{minipage}
\hspace{5pt}
\begin{minipage}{.28\textwidth}
    \centering
    \includegraphics[width=1.0\linewidth,trim={7cm, 6.5cm, 5.5cm, 5cm},clip]{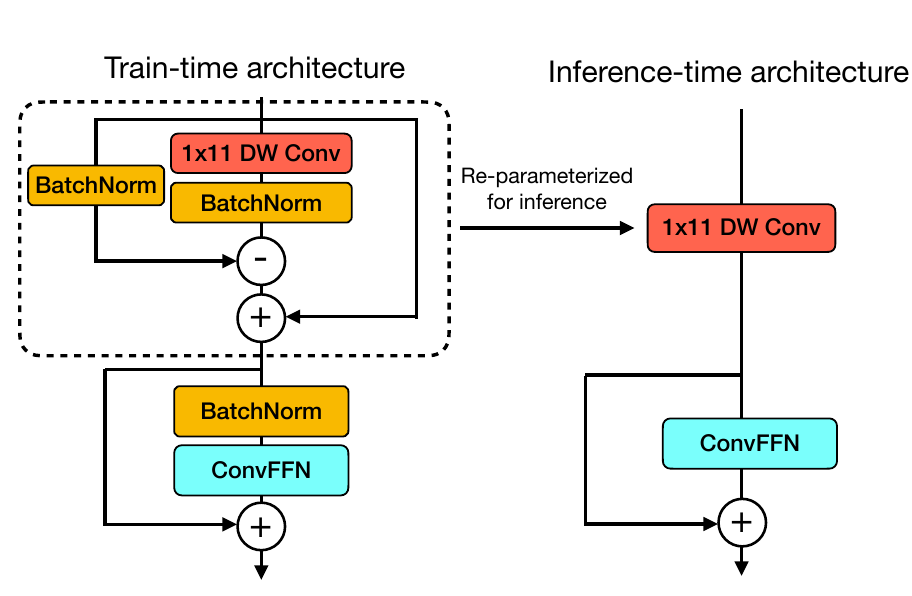}
    \captionof{figure}{Architecture of convolutional and reparameterizable blocks, 
    called \ourtextblock{} used in \ourmethod{}'s text encoder \ourtextzero{}.}
    \label{fig:text_repmixer_arch}
\end{minipage}
    \vspace{-5pt}
\end{figure*}

Our multi-modal reinforced training leverages knowledge transfer from an image captioning model and a strong ensemble of pretrained CLIP models for training the target model. It consists of two main components: i) leveraging the knowledge of an image captioning model via synthetic captions, and ii) knowledge distillation of image-text alignments from an ensemble of strong pre-trained CLIP models. We follow the dataset reinforcement strategy of~\citep{DR} and store the additional knowledge (synthetic captions and teacher embeddings) in the dataset (see \cref{fig:dr}), thereby avoiding any additional training time computational overhead such as evaluating the captioning model or the ensemble teacher. 

\subsection{Dataset Reinforcement}
\begin{figure}[t!]
    \centering
    \includegraphics[width=0.99\linewidth]{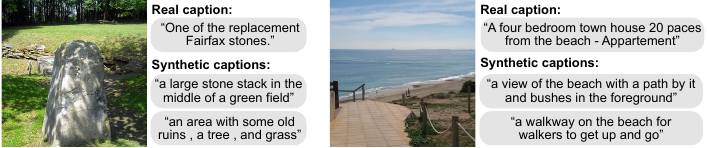}
    \caption{Real vs synthetic captions.}
    \label{fig:captioning}
\vspace{-10pt}
\end{figure}

\paragraph{Synthetic captions.} Image-text datasets used to train CLIP models 
are mostly sourced from the web, which is inherently noisy. Recent efforts 
such as \datacomp{}~\cite{DataComp} and data filtering networks~\citep{DFN} 
improve the quality of web-sourced datasets by using extensive filtering 
mechanisms. While these filtered datasets have lower noise, the captions may 
still not be descriptive enough. In order to boost the visual descriptiveness 
of the captions we use the popular CoCa~\citep{COCA} model and generate 
multiple synthetic captions $\syntextis$ for each image $\imagei$ 
(see \cref{fig:dr}a). Ablations on the number of synthetic captions generated 
per image are provided in \cref{sec:ablation}. \Cref{fig:captioning} shows some 
examples of synthetic captions generated by the CoCa model. Real captions in comparison to synthetic captions are generally more specific but noisier. We show (\cref{tab:real_vs_synth_ablation}) a combination of both real and synthetic captions is crucial to obtain best zero-shot retrieval and classification performance.

\vspace{-10pt}
\paragraph{Image augmentations.} For each image $\imagei$, we generate multiple 
augmented images $\augimageij$ using a parametrized augmentation function 
$\mathcal{A}$:
\begin{equation}
    \augimageij=\mathcal{A}(\imagei; \paramsij)\, ,
\end{equation}
where $\paramsij$ are the augmentation parameters that are sufficient to reproduce $\augimageij$ from $\imagei$ (see \cref{fig:dr}a). Ablations on the number and different kinds of augmentations used per image are provided in \cref{tab:num_aug_ablation,tab:teacher_aug_ablation}, respectively.

\vspace{-10pt}
\paragraph{Ensemble teacher.} Model ensembling is a widely used technique for 
creating a stronger model from a set of independently trained 
ones~\citep{DeepEnsemble,XCL}.
We extend this technique to multi-modal setup and use an ensemble of $K$ CLIP models as a strong teacher
(see \cref{sec:ablation} for our teacher ablations).
We compute the feature embeddings of these models for 
augmented images $\augimageij$ and synthetic captions $\syntextis$ obtaining 
$d_k$-dimensional vectors $\FeatTeacherImgijk$ and $\FeatTeacherSynisk$ for the 
$k$-th teacher model. We also compute the teacher embeddings $\FeatTeacherTxtik$ of the 
ground-truth captions $\gttexti$ (see \cref{fig:dr}b).

\vspace{-10pt}
\paragraph{Reinforced dataset.} We store the image augmentation parameters 
$\paramsij$, synthetic captions $\syntextis$, feature embeddings 
$\FeatTeacherImgijk$, $\FeatTeacherSynisk$ and $\FeatTeacherTxtik$ of the CLIP 
teachers as additional knowledge in the dataset along with the original image 
$\imagei$ and caption $\gttexti$ (see \cref{fig:dr}c). Note that dataset reinforcement is a one-time cost that is amortized by several efficient model training and experimentation.

\subsection{Training}
\paragraph{Loss function.}
Intuitively, our loss function distills the affinity matrix
between image-text pairs from multiple image-text teacher encoders
into student image-text encoders.
Let $\mathcal{B}$ denote a batch of $b$ (image, 
text) pairs and $\FeatTeacherImgk, \FeatTeacherTxtk \in \mathcal{R}^{b \times 
d_k}$ the matrices of $d_k$-dimensional image and text embeddings, 
respectively, of the $k$-th model in the teacher ensemble for batch 
$\mathcal{B}$. Correspondingly, we denote the image and text embedding matrices of the target model by $\FeatStudentImg, \FeatStudentTxt \in \mathcal{R}^{b \times 
d}$.  
For given $\UU$ and $\VV$ matrices, let $\Similarity_\tau(\UU, \VV) \in \mathcal{R}^{b 
\times b}$ denote their similarity matrix obtained by applying row-wise Softmax operation to $\UU\VV^{\top}/\tau$, where
$\tau$ is a temperature parameter.
Our training loss consists of two 
components, the standard CLIP~\citep{CLIP} loss $\LClip(\mathcal{B})$ and 
a knowledge distillation loss $\LKD(\mathcal{B})$:
\begin{align}\label{eq:distill}
\LTotal(\mathcal{B}) &= (1-\lambda) \LClip(\mathcal{B}) + \lambda \LKD(\mathcal{B}),\\
    \LKD(\mathcal{B}) &=
    \frac{1}{2}\LKDIT(\mathcal{B})
    + \frac{1}{2} \LKDTI(\mathcal{B}),\notag\\
    \LKDIT(\mathcal{B}) &= \frac{1}{bK} \sum_{k=1}^{K}
    \text{KL}(
    \Similarity_{\tau_k}(\FeatTeacherImgk, \FeatTeacherTxtk)
    \|
    \Similarity_{\widehat{\tau}}(\FeatStudentImg, \FeatStudentTxt))\notag,
\end{align}
where KL denotes Kullback-Leibler divergence, $\LKDTI$ is computed by swapping the text and image embedding terms of $\LKDIT$, and $\lambda$ is a tradeoff parameter.

\paragraph{Efficient training.} Training on the reinforced dataset is as simple as modifying the data loader and loss function to exploit additional knowledge stored in the 
dataset and has the same training cost as standard CLIP training (see \cref{tab:training_time}). For every sample, we read the image $\imagei$ and the corresponding 
ground-truth caption $\gttexti$ from the dataset. Then, we randomly load one of 
stored augmentation parameters $\paramsij$ and reproduce the augmented image 
$\augimageij$. We also randomly load one of synthetic captions $\syntextis$.  
Finally, we read the stored embeddings, $\FeatTeacherImgijk$, 
$\FeatTeacherSynisk$, and $\FeatTeacherTxtik$, corresponding to the $K$ teacher 
models. 

Using this loaded data, we construct two data batches, $\mathcal{B}_{\text{real}}$ 
corresponding to (augmented image, real caption) pairs and $\mathcal{B}_{\text{syn}}$ 
corresponding to (augmented image, synthetic caption) pairs, and compute our training loss in \cref{eq:distill} separately on  $\mathcal{B}_{\text{real}}$ and
$\mathcal{B}_{\text{syn}}$. Our final loss is given by
\begin{equation}\label{eq:final_total_loss}
\sum_{\mathcal{B} \in \{B_{\text{real}}, B_{\text{syn}} \} } \LTotal(\mathcal{B}).
\end{equation}
Note that we can compute the total loss after a forward pass of the student model without any extra teacher-related computations since the teacher embeddings required to compute the distillation loss are readily available as part of the dataset.

\section{Architecture}

\subsection{Text Encoder}
\vspace{-3pt}

CLIP~\cite{CLIP} model paired the vision transformer with a classical transformer comprising of self-attention layers for text encoding. While this model is effective, smaller and more efficient models are preferred for mobile deployment. Recently, works like~\cite{lightconv} have shown that convolutions can be just as effective for text encoding. In contrast, we found that purely convolutional architectures significantly underperform their transformer counterparts. Instead of using a fully convolutional architecture for text encoding, we introduce a hybrid text encoder which makes use of 1-D convolutions and self-attention layers. 

For hybrid text encoder, we introduce \textit{\ourtextblock{}}, a convolutional 
token mixer that decouples train-time and inference-time architectures.  
\ourtextblock{} is inspired by reparameterizable convolutional token mixing (RepMixer) introduced in~\citep{vasu2023fastvit}. At inference, skip 
connections are reparameterized. The  architecture is shown in \cref{fig:text_repmixer_arch}. For Feed-Forward Network (FFN) blocks, we augment linear layers with an additional depthwise 1-D convolution of similar kernel dimensions as the token mixer, to obtain \textit{ConvFFN} blocks. This structure is similar to the convolutional blocks used in~\citep{conformer}, the main difference being the use of batchnorm and the ability to fold it with the succeeding depthwise 1-D convolutional layer for efficient inference. 
The design choices for \textit{\ourtextblock{}} is discussed in \cref{sec:hybrid_text_encoder}.
In order to find the optimal design for our hybrid text encoder, we start with a purely convolutional text 
encoder and start replacing convolutional blocks systematically with 
self-attention layers (see \cref{tab:hybrid_text}). \cref{tab:mci_mct}, show the efficacy of our text encoder when compared with CLIP's base text encoder. Our model is smaller, faster and obtains similar performance as the larger base text encoder when paired with efficient backbones like ViT-S/16.

\subsection{Image Encoder}\label{sec:image_encoder}
Recent works have shown the efficacy of hybrid vision transformer for learning good visual representations. For \ourmethod{}, we introduce an improved hybrid vision transformer called \ourimage{} based on the recent FastViT~\cite{vasu2023fastvit} architecture with certain key differences explained below.

In FastViT, an MLP expansion ratio of 4.0 is used for FFN blocks. Recent works like~\cite{efficientvit, tinyclip} exposed the significant amount of redundancy in linear layers of FFN block. To improve parameter efficiency, we simply lower the expansion ratio to 3.0 and increase the depth of the architecture. By doing so, we retain the same number of parameters in the image encoder. The stage configuration for the three variants are described in \cref{sec:mci_configs}. \ourimagezero{} has similar stage configuration as~\cite{MobileOne}. \ourimageone{}, is a deeper version of \ourimagezero{} and \ourimagetwo{} is a wider version of \ourimageone{}. The stage compute ratios in our variants are similar to~\cite{MobileOne}. We find that this design has a minimal impact on latency, but a good improvement in capacity of the model, reflected in the downstream task performance, see \cref{sec:supp_exp_setup}. In~\cref{tab:mci_mct}, we compare our MCi encoder with a similar sized FastViT-MA36 when used as image encoders in a CLIP model. Our model obtains much better zero-shot IN-val performance while being 16.3\% faster.

\begin{table}[h!]
    \centering
    \resizebox{0.45\linewidth}{!}{
    \begin{subtable}[t]{.65\linewidth}
    \begin{tabular}{c|C{1cm}|C{1cm}}
            \toprule[1.5pt]
            \textbf{Text Enc.}
            & \textbf{Latency {(txt)}}
            & \textbf{0-shot \imagenetval{}}
            \\
             \midrule[1.5pt]
              Base       & 3.3 & 53.4 \\
              \midrule[0.5pt]
              \ourtext{} (Ours) & \textbf{1.6} & \textbf{53.6} 
              \\
            \bottomrule[1.5pt]
        \end{tabular}
    \end{subtable}
    }
    \resizebox{0.45\linewidth}{!}{
    \begin{subtable}[t]{.65\linewidth}
    \begin{tabular}{c|  C{1cm}|C{1cm}}
            \toprule[1.5pt]
            \textbf{Image Enc.}
            & \textbf{Latency {(img)}}
            & \textbf{0-shot \imagenetval{}}
            \\
             \midrule[1.5pt]
            FastViT-MA36   & 4.3  & 58.9 \\
            \midrule[0.5pt]
            \ourimagetwo{} (Ours) & \textbf{3.6} & \textbf{60.0} \\
 
            \bottomrule[1.5pt]
        \end{tabular}

    \end{subtable}
    }%
     \vspace{-6pt}
    \caption{\textbf{(a)} \textbf{Base vs.\ \ourtext{}} text encoders with ViT-S/16. \textbf{(b) FastViT vs.\ \ourimage{}} image encoders with Base text encoder. Trained for 30k iters ($\sim$0.24B seen samples) on \ourdatasetTM{}.}\label{tab:mci_mct}
     \vspace{-10pt}
\end{table}

\section{Experiments}\label{sec:experiments}

In this section, we present our experimental setup and results.
\vspace*{-10pt}
\paragraph{Evaluation.}

We evaluate image-text models using the evaluation benchmark of 
DataComp~\citep{DataComp}. Specifically, we report zero-shot classification on 
the ImageNet validation set~\citep{IN-1k}, and its distribution shifts 
including ImageNet-V2~\citep{IN-v2}, ImageNet-A~\citep{IN-AO}, 
ImageNet-O~\citep{IN-AO}, ImageNet-R~\citep{IN-R}, and 
ObjectNet~\citep{ObjectNet}, which we report their average as IN-Shift. For zero-shot image-text retrieval, we 
report recall@1 on MSCOCO~\citep{MSCOCO} and Flickr30k~\citep{Flickr30k} 
datasets.  Further, we report average performance on all 38 datasets in 
DataComp evaluations.  We also evaluate our models on Visual Genome Relation, 
Visual Genome Attributes, Flickr30k-Order and COCO-Order datasets which are part 
of the recent Attribute, Relation and Order (ARO) benchmark~\citep{vlmbow}. In the remainder, 
\imagenetval{} refers to zero-shot accuracy on ImageNet validation 
set and \flickrval{} refers to average zero-shot recall@1 for image-text and 
text-image retrieval. All reported metrics are obtained without any fine-tuning.

\vspace*{-10pt}
\paragraph{Training setup.}

We have two setups for ablations and large-scale experiments. For 
ablations, we train on datasets with 12.8M image-text pairs using a global 
batch size of 8,192 and 8$\times$NVIDIA-A100-80GB GPUs for 30-45k 
iterations.  For large-scale training, we use a global batch size of 65,536 
with 256$\times$A100 GPUs for 200k iterations. All models are trained from 
scratch (see details in \cref{sec:supp_exp_setup}).

\vspace*{-10pt}
\paragraph{Dataset.}
We train on the image-text dataset of DataComp dataset~\citep{DataComp}. We use 
the Bestpool filtered subset of 1.28B samples that provides them with best 
performance at the largest dataset scale. We refer to this set as 
\datacompOneB{}. For fast experimentation, we create a fixed subset of 12.8M 
uniformly sampled pairs which we call \datacompTM{}.  \datacompTM{} was not 
studied in \citep{DataComp} but in our experiments, we observed that 
\datacompTM{} consistently achieves better performance compared with the 
Bestpool subset of DataComp-medium with comparable samples.

\vspace*{-10pt}
\paragraph{DataCompDR: Reinforced DataComp.}
We reinforce the DataComp dataset using our multi-modal dataset reinforcement 
strategy. In particular, we create \ourdatasetOneB{} and \ourdatasetTM{} by 
reinforcing \datacompOneB{} and \ourdatasetTM{}. We have a one-time 
generation process, the cost of which is amortized over multiple architectures 
and extensive ablations. We generate 5 synthetic captions per image using the 
\verb|coca_ViT-L-14| model in OpenCLIP~\citep{OpenCLIP}, and strong random 
image augmentations (10 for \ourdatasetOneB{} and 30 for \ourdatasetTM{}). We 
compute embeddings of an ensemble of two strong teachers 
(\verb|ViT-L-14| with pretrained weights \verb|datacomp_xl_s13b_b90k| and 
\verb|openai| in OpenCLIP) on augmented images as well as real and synthetic 
captions.
Embeddings are 1536-D concatenations of 2$\times$768-D vectors.
We store all reinforcements using lossless compression and BFloat16.
We analyze all of our choices in \cref{sec:ablation}.
One seen sample
for \ourdataset{} is a triplet of one randomly augmented image, one ground-truth caption, and one randomly picked synthetic caption.

\vspace*{-10pt}
\paragraph{\ourmodel{} architectures.}
Our \ourmodel{} architectures are formed as pairs of \ourimage{}:\ourtext{} 
architectures. In particular, we create 3 small variants \ourmodelSZero{} 
(\ourimagezero{}:\ourtextzero{}), \ourmodelSOne{} (\ourimageone{}:Base), and 
\ourmodelSTwo{} (\ourimagetwo{}:Base), where Base is a 12-layer Transformer 
similar to the text-encoder of ViT-B/16 based CLIP~\citep{CLIP}. We also train 
a standard pair of ViT-B/16:Base and refer to our trained model as \ourmodelB{}.

\vspace*{-10pt}
\paragraph{Benchmarking latency.}
To measure latency, we use the input sizes corresponding to the respective 
methods. For iPhone latency measurements, we export the models using Core ML 
Tools (v7.0)~\citep{coremltools} and run it on iPhone12 Pro Max with iOS 
{17.0.3}. Batch size is set to 1 for all the models. We follow the same 
protocol as described in~\cite{MobileOne}. 

\subsection{Ablation Studies}\label{sec:ablation}
\begin{table}[t!]
        \centering
    \resizebox{0.8\columnwidth}{!}{
    \begin{tabular}{C{0.7cm}C{1.4cm}C{1.2cm}C{1cm}|C{1.2cm}C{1.2cm}}
        \toprule[1.5pt]
        $\lambda$&
        Syn. Captions& 
        Strong Aug.&
        Ens. Teacher&
        \imagenetval{}&
        \flickrval{}\\
        \midrule[1.25pt]
        0 & \xmark & \xmark & \xmark & 44.5 & 41.8\\
        0 & \cmark & \xmark & \xmark & 51.9 & 69.3\\
        1 & \cmark & \xmark & \xmark & 54.5 & 66.1\\
        1 & \cmark & \cmark & \xmark & 59.3 & 70.5\\
        \rowcolor{blue!25}1 & \cmark & \cmark & \cmark & \underline{61.7} & 72.0\\
        \rowcolor{gray!25}0.7 & \cmark & \cmark & \cmark & 60.7 & \underline{74.2}\\
        \bottomrule[1.5pt]
    \end{tabular}}

    \caption{\textbf{Summary of ablations.} We train on \ourdatasetTM{} for 30k 
    iterations.  All ablations are on ViT-B/16:Base. We highlight our main 
    choices with {\colorbox{blue!25}{blue}} and alternative tradeoffs with 
    {\colorbox{gray!25}{gray}}. We \underline{underline} numbers within $0.5\%$ 
    of the maximum.}
    \label{tab:method_ablation}
    \vspace{-10pt}
\end{table}
In this section, we analyze the effect of each component in our training and architecture.
Unless otherwise stated, we use ViT-B/16:Base encoders trained on \datacompTM{} for 30k iterations 
with global batch size of 8k ($\sim$20 epochs).
\Cref{tab:method_ablation} summarizes the analysis of our training.

\vspace*{-10pt}
\paragraph{Strong image augmentations.}
In contrast to uni-modal supervised and self-supervised methods for vision 
with strong augmentations~\citep{dosovitskiy2020image,touvron2021training}, CLIP training 
recipes~\citep{CLIP} often use light image augmentations to avoid image-text 
mis-alignment. However, several works~\citep{XCL,DR,beyer2022knowledge} 
demonstrated the efficacy of strong augmentations in a distillation setup. In 
\cref{tab:method_ablation} we show that strong image augmentations improve 
distillation performance (+4.8\% on \imagenetval{} and +4.4\% on \flickrval{}).  We 
provide detailed ablation on the effect of image augmentations in 
\cref{sec:augmentation_ablation}.

\vspace*{-10pt}
\paragraph{Synthetic captions.}
Similar to image augmentations, synthetic captions (or caption 
augmentations) can further improve the performance of CLIP models,
particularly on image-text retrieval. For regular CLIP training ($\lambda=0$), we observe in \cref{tab:method_ablation} that including batches with both synthetic and real captions results in +7.4\% on \imagenetval{} and +27.5\% on \flickrval{} performance improvements. In \cref{tab:real_vs_synth_ablation}, we observe a similar trend for CLIP training with distillation loss only ($\lambda=1$).
In 
\cref{tab:lambda_ablation}, we analyze the effect of $\lambda$ and observe 
a tradeoff where $\lambda=1.0$ is optimal for 
\imagenetval{} while $\lambda=0.7$ is optimal for \flickrval{}.
Prior work that exploit synthetic captions primarily focus on improved 
retrieval~\citep{VeCLIP,ALIP} while distillation works focus on zero-shot 
classification~\citep{DIME-FM}. In our large-scale experiments, we balance the 
tradeoff for \ourmodelB{} using $\lambda=0.75$ and use $\lambda=1.0$ for our 
small variants.

\vspace*{-10pt}
\paragraph{Ensemble teacher.}
We find that using an ensemble of strong CLIP models as a teacher in 
our multi-modal reinforced training is crucial to achieving +2.4\% 
\imagenetval{} improvement (\cref{tab:method_ablation}). We also observe that the most 
accurate models are not the best teachers.  See \cref{sec:clip_ensembles} for 
a comprehensive analysis of different teacher models.

\begin{table}[t!]
\centering
\begin{subtable}{0.9\columnwidth}
        \centering
    \resizebox{0.9\columnwidth}{!}{
        \begin{tabular}{c|cccc}
            \toprule[1.5pt]
            $\mathcal{B} \in$&
            $\{ \mathcal{B}_{\text{real}}\}$ &
            \cellcolor{gray!25}$\{ \mathcal{B}_{\text{syn}}\}$ &
            $\{\mathcal{B}_{\text{real}}$ or $\mathcal{B}_{\text{syn}}\}$ &
            \cellcolor{blue!25}
            $\{\mathcal{B}_{\text{real}}$, $\mathcal{B}_{\text{syn}}\}$
            \\
            \midrule
            \imagenetval{} &
            56.4& 49.8 & 57.3& \underline{61.7}\\
            \flickrval{}&
            57.0& \underline{72.2}& 68.6 &\underline{72.0} \\
            \bottomrule[1.5pt]
        \end{tabular}
    }

    \caption{Real vs synthetic sampling in \cref{eq:final_total_loss} ($\lambda=1.0$).}
    \label{tab:real_vs_synth_ablation}
\end{subtable}
\newline
\vspace*{-8pt}
\newline
\begin{subtable}{0.99\columnwidth}
    
  \centering
  \resizebox{0.99\columnwidth}{!}{
  \begin{tabular}{c|cccccccccc}
    \toprule[1.5pt]
    $\lambda$& 0.1& 0.2& 0.3& 0.4& 0.5& 0.6& \cellcolor{blue!25}0.7& 0.8& 0.9& 1.0\\
    \midrule
    \imagenetval{}& 54.4& 56.3& 57.4& 58.2& 59.5& 60.3& 60.7& \underline{61.5}& \underline{61.6}& \underline{61.7}\\
    \flickrval{}& 71.4& 71.5& 71.8& 72.2& \underline{73.8}& 73.6& \underline{74.2}& 73.1& 73.2& 72.0\\
    \bottomrule[1.5pt]
  \end{tabular}}

    \caption{Ablation on the loss coefficient ($\lambda$) in 
    \cref{eq:distill}.}
    \label{tab:lambda_ablation}
\end{subtable}
\vspace{-5pt}
    \caption{\textbf{Ablation on the loss.} The tradeoff between \imagenetval{} 
    and \flickrval{} is controlled by the synthetic sampling and loss coefficient. We 
    train for 30k iterations.}
\vspace{-5pt}
\end{table}

\begin{table}[t!]
    \centering
\begin{subtable}{0.95\columnwidth}
    
    \centering
    \resizebox{0.99\columnwidth}{!}{
        \begin{tabular}{c|cccccccc}
            \toprule[1.5pt]
            Num. Aug. &
            1    & 2    & \cellcolor{gray!25}5    & 10   & 15   & 20   & 25   & \cellcolor{blue!25}30\\
            \midrule
            \imagenetval{} &
            60.63 & 63.27 & \underline{64.81} & \underline{64.74} & \underline{64.49} & \underline{64.92} & \underline{64.78} & \underline{64.74}\\
            \flickrval{} &
            69.61& 71.74  & 74.76 & 74.46 & 73.90 & 74.29 & 73.27 & \underline{75.66}\\
            \bottomrule[1.5pt]
        \end{tabular}
    }

    \caption{Effect of the number of augmentations.}
    \label{tab:num_aug_ablation}
\end{subtable}
\newline
\vspace*{-8pt}
\newline
\begin{subtable}{0.85\columnwidth}
    
    \centering
    \resizebox{0.90\columnwidth}{!}{
        \begin{tabular}{c|cccccc}
            \toprule[1.5pt]
            Num. Caps. &
            0 & 1 & \cellcolor{gray!25}2 & 3 & 4 & \cellcolor{blue!25}5\\
            \midrule
            \imagenetval{} &
            60.67 & \underline{64.88} & \underline{65.19} & \underline{65.19} & \underline{64.81} & \underline{64.74}\\
            \flickrval{} &
            62.26 & 73.82 & 74.27 & 73.91 & 74.07 & \underline{75.66} \\
            \bottomrule[1.5pt]
        \end{tabular}
    }

    \caption{Effect of the number of synthetic captions.}
    \label{tab:num_cap_ablation}
\end{subtable}
\newline
\vspace*{-8pt}
\newline
\begin{subtable}{0.9\columnwidth}
    
    \centering
    \resizebox{0.99\columnwidth}{!}{
        \begin{tabular}{ccccC{1cm}C{1cm}C{1cm}|C{0.7cm}}
            \toprule[1.5pt]
            Dataset        & Image  & Text   & Syn.   & Aug. Params & Text Emb. & Image Emb.& Size (TBs)\\
            \midrule[1.25pt]
            \datacompTM{}   & \cmark & \cmark & \xmark & \xmark      & \xmark    & \xmark    & 0.9\\
                           & \cmark & \cmark & \cmark & \cmark      & \xmark    & \xmark    & 0.9\\
            \cellcolor{blue!25}\ourdatasetTM  & \cmark & \cmark & \cmark & \cmark      & 5+1       & 30        & 1.9\\
            \midrule
            \datacompOneB{}    & \cmark & \cmark & \xmark & \xmark      & \xmark    & \xmark    & 90\\
            \cellcolor{blue!25}\ourdatasetOneB& \cmark & \cmark & \cmark & \cmark      & 5+1       & 10        &140\\
            \bottomrule[1.5pt]
        \end{tabular}
    }

    \caption{Total storage for samples stored in individual Pickle Gzip files 
    and BFloat16 embeddings.  $+1$ refers to the ground-truth caption. For 
    further size reductions see \cref{tab:storage_size_full}.}
    \label{tab:storage_size}
\end{subtable}
\newline
\vspace*{-8pt}
\newline
\begin{subtable}{0.9\columnwidth}
    
    \centering
    \resizebox{0.99\columnwidth}{!}{
        \begin{tabular}{C{2.8cm}C{2cm}C{0.8cm}C{0.8cm}C{1.8cm}C{1.8cm}|C{1cm}}
            \toprule[1.5pt]
            Dataset&
            $\mathcal{B}\in$&
            $\LClip$&
            $\LKD$&
            Stored Syn.  Caption &
            Stored Embeddings &
            Time (hours)
            \\
            \midrule[1.25pt]
            \datacompTM{}& $\{\mathcal{B}_{\text{real}}\}$ & \cmark & \xmark &
           \xmark & \xmark &    \underline{1.3}\\
            \midrule
            -& $\{\mathcal{B}_{\text{real}}$, $\mathcal{B}_{\text{syn}}\}$ & \cmark & \cmark & \xmark & \xmark &    21.1\\
             - & $\{\mathcal{B}_{\text{real}}$, $\mathcal{B}_{\text{syn}}\}$  & \cmark & \cmark & \cmark & \xmark &  4.1\\
             \cellcolor{blue!25}\ourdatasetTM{}& $\{\mathcal{B}_{\text{real}}$, $\mathcal{B}_{\text{syn}}\}$ & \cmark & \cmark & \cmark & \cmark &  \underline{1.3}\\
            \bottomrule[1.5pt]
        \end{tabular}
    }

    \caption{Training time per epoch (12.8M samples) on $8\times$A100-80GB.}
    \label{tab:training_time}
\end{subtable}
\vspace{-5pt}
    \caption{\textbf{Ablations on storage/cost.} Training on \ourdataset{} has 
    no time overhead. We train for 45k iterations ($\sim$30 epochs).}
\vspace{5pt}
\end{table}

\begin{table}[t!]
\centering
\vspace*{-8pt}
\begin{subtable}{0.67\columnwidth}

    \centering
    \resizebox{0.99\columnwidth}{!}{
        \begin{tabular}{c|ccccc}    %
            \toprule[1.5pt]
            Num. Self-attn. & 6 & \cellcolor{blue!25}4 & 2 & 1 & 0 \\
            \midrule
            Num Params. (M) & 44.5 & 42.4 & 40.4 & 39.3 & 38.3 \\
            Latency (ms) & 1.9 & 1.6 & 1.4 & 1.3 & 1.2 \\
            \midrule
            \imagenetval{} & \underline{60.9} & \underline{60.8} & 60.2 & 60.0 & 57.9 \\
            \bottomrule[1.5pt]
        \end{tabular}
    }

\end{subtable}
\vspace{-5pt}
    \caption{\textbf{Ablation on architecture.} Effect of the number of self-attention layers in \ourtext{}. We train for 30k iterations.}\label{tab:hybrid_text}
\vspace{-7pt}
\end{table}

\vspace*{-10pt}
\paragraph{Number of image augmentations and synthetic captions.}
We generate multiple image augmentations and synthetic captions and store them 
efficiently along with the teacher embeddings. We investigate the effectiveness 
of the number of augmentations and synthetic captions in 
\cref{tab:num_aug_ablation,tab:num_cap_ablation}. We train models with up to 30 
image augmentations and 5 synthetic captions for 45k iterations ($\sim$30 
epochs).  We observe that the performance nearly saturates at {5} augmentations 
and {2} synthetic captions suggesting each augmentation can be reused 
multiple times before the added knowledge is fully learned by the model.
When needed, fewer augmentations and synthetic captions can help reduce the generation time 
and storage overhead.
For maximal performance, we reinforce \ourdatasetTM{} and \ourdatasetOneB{} with 10 and 30 augmentations, respectively,
and 5 synthetic captions.

\vspace*{-10pt}
\paragraph{Training time.}
A major advantage of reinforced training is the minimal time difference with 
non-reinforced training.
We provide the wall-clock times in \cref{tab:training_time} for
regular CLIP training as well as training with online distillation
and a caption generator.  We measure the time for 
training on one epoch of \ourdatasetTM{} on a single node with 8$\times$ 
A100-80GB GPUs.  An epoch takes {1562} iterations with global batch size 8192 
on \ourdatasetTM{}. Without any dataset reinforcement, training is 16$\times$ 
slower while with partial reinforcements of synthetic captions it is 3$\times$ 
slower.

\vspace*{-10pt}
\paragraph{Storage size.}
We report the storage requirements for our reinforced datasets compared with 
the original DataComp dataset. 
We report the storage size of one file per image-text pair. If present, 
we store all corresponding reinforcements in the same file.  We store files in 
the Pickle format and compress each file with Gzip compression. The
image-text embeddings are saved in BFloat16.
We report the total storage size for 12.8M samples of \ourdatasetTM{} and 
{1.28B} samples of \ourdatasetOneB{} in \cref{tab:storage_size}.
We provide analysis on additional size reductions in \cref{sec:ablations_lossy} 
and verify that using BFloat16 does not impact the accuracy.
For minimal storage overhead, we recommend {5} augmentations/synthetic captions 
for 30 epochs on \ourdatasetTM{} and {2} for 10 epochs on \ourdatasetOneB{} 
which are based on our ablations in 
\cref{tab:num_aug_ablation,tab:num_cap_ablation}.

\vspace*{-10pt}
\paragraph{Hybrid text encoder.}
We ablate over the number of \ourtextblock{} blocks that can effectively replace 
self-attention layers with negligible impact on zero-shot performance. For this 
ablation, we choose a 6-layer purely convolutional text encoder and 
systematically introduce self-attention layers in the middle. From 
\cref{tab:hybrid_text}, we find that even introducing a single self-attention 
layer substantially improves the zero-shot performance. The best tradeoff 
is with 2 blocks of \ourtextblock{} and 4 blocks of self-attention 
layers.  This variant, \ourtextzero{}, obtains similar performance as the pure 
transformer variant, while being 5\% smaller and 15.8\% faster.

\subsection{Small Scale Regime}\label{sec:dr12m}

In \cref{tab:small_scale_clip}, we compare methods trained on datasets with 
12-20M samples, a relatively small range for fast exploration (e.g., architecture search). \ourmodelB{} trained on \ourdatasetTM{} with less than 370M samples significantly outperforms all other methods with up to 4$\times$ longer training. Also \ourmodelB{} shows great scaling with number of seen samples (65.3$\to$71.7\%) in comparison to previous work SLIP~\citep{SLIP}(42.8$\to$45.0\%).  In comparison to CLIPA~\citep{CLIPA} which uses multi-resolution training for efficiency, training with \ourdatasetTM{} is more efficient: CLIPA obtains 63.2\% with 2.69B multi-resolution seen samples (which has equivalent compute as $\sim$0.5B $224^2$ seen samples), that is worse than \ourmodelB{}'s 65.3\% with only 0.37B seen samples. Further, TinyCLIP-39M/16 in comparison to \ourmodelSTwo{} has higher latency and less accuracy, and TinyCLIP-8M/16 is significantly less accurate than \ourmodelSZero{} (41.1\% vs 59.1\%) while having a close latency (2.6 ms vs 3.1 ms).

\begin{table}[t!]
    \centering
    \resizebox{0.99\columnwidth}{!}{
        \begin{tabular}{l|cC{1.5cm}|C{2cm}|C{1.6cm}}
            \toprule[1.5pt]
            \textbf{Name} &
            \textbf{Dataset}
            & \textbf{Seen Samples}
            & \textbf{Latency (ms) {(img+txt)}}
            & \textbf{Zero-shot \imagenetval{}}
            \\
             \midrule[1.25pt]

            CLIP-B/16~\citep{CLIP,SLIP}&
            CC-12M~\citep{CC12M}&
            0.39B&
            \multirow{2}{*}{11.5 + 3.3}& 
            36.5\\
            
            CLIP-B/16~\citep{CLIP,SLIP}&
            YFCC-15M~\citep{yfcc}&
            0.37B&
            &
            37.6\\

            \textbf{\ourmodelB{}} &
            CC-12M~\citep{CC12M}&
            0.37B &
            10.4 + 3.3& 
            38.1\\

            SLIP-B/16~\citep{SLIP}&
            CC-12M~\citep{CC12M}&
            0.39B &
            \multirow{2}{*}{11.5 + 3.3} &
            40.7\\
            
            SLIP-B/16~\citep{SLIP}&
            YFCC-15M~\citep{yfcc}&
            0.37B&
            & 
            42.8\\

            \textbf{\ourmodelB{}} &
            \datacompTM{}~\citep{DataComp}&
            0.37B &
            10.4 + 3.3& 
            50.1\\
            
             \textbf{\ourmodelB{}}&
             \ourdatasetTM{}&
             0.37B&
             10.4 + 3.3&
             \textbf{65.3}\\

            \midrule

            CLIP-B/32~\citep{CLIP,DeFILIP}&
            \multirow{5}{*}{YFCC-15M~\citep{yfcc}}&
            \multirow{5}{*}{0.49B} & 
            \multirow{5}{*}{5.9 + 3.3}& 
            32.8\\

            SLIP-B/32~\citep{SLIP,DeFILIP}&
            & 
            &
            & 
            34.3\\

            FILIP-B/32~\citep{FILIP,DeFILIP}&
            &
            & 
            &
            39.5\\

            DeCLIP-B/32~\citep{DeCLIP}&
            &
            & 
            &
            43.2\\

            DeFILIP-B/32~\citep{DeFILIP}&
            &
            & 
            &
            45.0\\

            RILS-B/16~\citep{RILS}&
            LAION-20M~\citep{Laion400}&
            0.5B & 
            11.5 + 3.3& 
             45.0\\

            \midrule

            TinyCLIP-8M/16~\citep{tinyclip}&
            YFCC-15M~\citep{yfcc}& 
            0.75B &
            \textbf{2.0 + 0.6} & 
            41.1\\
            
            SLIP-B/16~\citep{SLIP}&
            YFCC-15M~\citep{yfcc}&
            0.75B & 
            11.5 + 3.3& 
            44.1\\

            CLIP-B/16 &
            \datacompTM{}~\citep{DataComp} &
            0.74B & 
            10.4 + 3.3 &
            53.5\\

            \textbf{\ourmodelSZero{}} &
             \ourdatasetTM{} &
             0.74B &
             \textbf{1.5 + 1.6} &
             \textbf{59.1}\\
            
            TinyCLIP-39M/16~\citep{tinyclip}&
            YFCC-15M~\citep{yfcc}&
            0.75B &
            5.2 + 1.9 & 
            63.5\\

             \textbf{\ourmodelSTwo{}} &
             \ourdatasetTM{} &
             0.74B &
             \textbf{3.6 + 3.3} &
             \textbf{64.6} \\
            
             \textbf{\ourmodelB{}} &
             \ourdatasetTM{} & 
             0.74B &
             10.4 + 3.3 &
             \textbf{69.1}\\

            \midrule

             SLIP-B/16~\citep{SLIP}&
            YFCC-15M~\citep{yfcc}&
            1.5B & 
            11.5 + 3.3& 
            45.0\\

             CLIP-B/16 & 
             \datacompTM{}~\citep{DataComp} &
             1.48B & 
             10.4 + 3.3 &
             55.7\\
             
             \textbf{\ourmodelB{}} &
             \ourdatasetTM{} &  
             1.48B & 
             10.4 + 3.3&
             \textbf{71.7}\\
             
             \midrule
             CLIPA-B/16~\citep{CLIPA} & 
             LAION-400M~\citep{Laion400} &
             2.69B$^\dagger$ & 
             11.5 + 3.3 &
             63.2\\
             
            \bottomrule[1.5pt]
        \end{tabular}
    }
    \vspace{-5pt}
    \caption{\textbf{Small-scale CLIP training.} \ourmodelB{} notation refers to our re-implementation of ViT-B/16 image encoder and standard Base text encoder. $^\dagger$ refers to multi-resolutions. Models are grouped based on the number of samples seen.}
    \label{tab:small_scale_clip}
    \vspace{-5pt}
\end{table}

\subsection{Learning Efficiency}\label{sec:learning_efficiency}
\begin{figure}
    \centering
    \begin{subfigure}[t]{0.49\linewidth}
        \includegraphics[width=\textwidth]{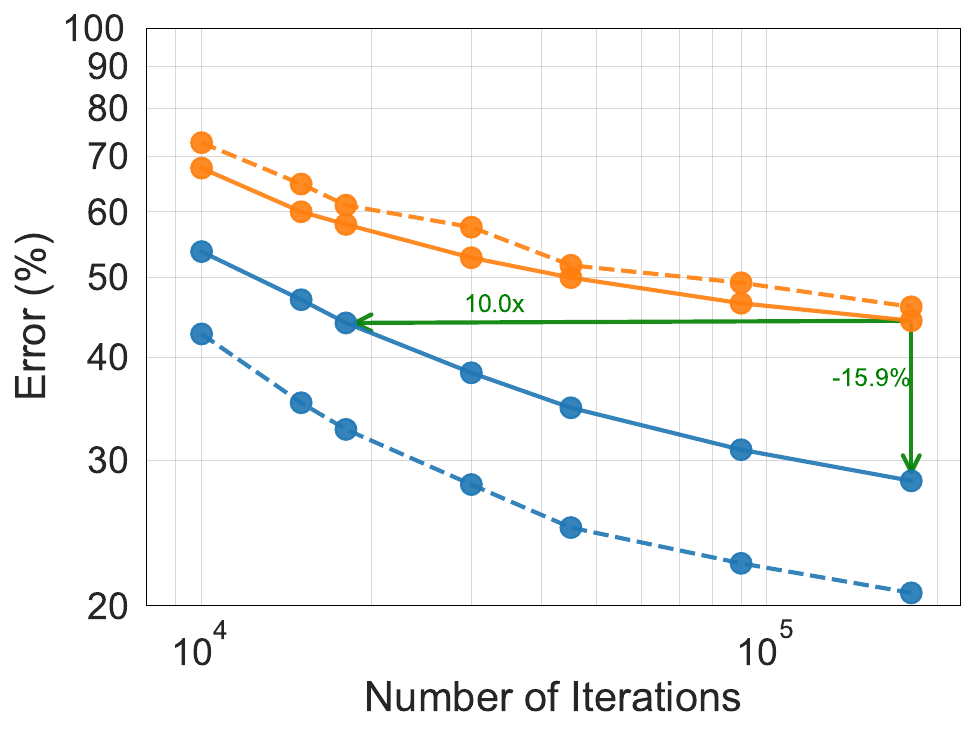}
        \caption{Iteration Scaling}
        \label{fig:scaling_iters}
    \end{subfigure}
    \hfill
    \begin{subfigure}[t]{0.49\linewidth}
        \includegraphics[width=\textwidth]{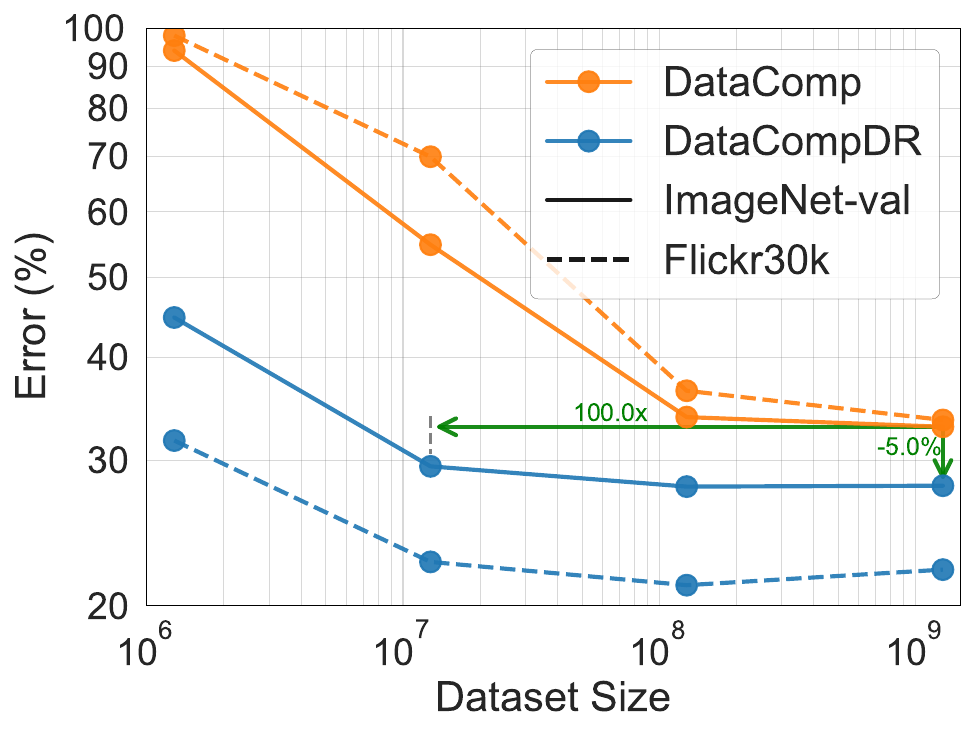}
        \caption{Data Scaling}
    \label{fig:scaling_data}
    \end{subfigure}
    \label{fig:scaling}
    \vspace{-5pt}
    \caption{\textbf{Learning efficiency up to 1000$\times$.} Training on \ourdataset{} is 
    10$\times$ more iteration efficient and 100$\times$ more data 
    efficient on \imagenet{}-val and 18$\times$ and 1000$\times$
    more efficient on \flickr{}
    compared with non-reinforced training.}
    \vspace{-5pt}
\end{figure}

Training longer with knowledge distillation is known to consistently improve performance
for classification models~\citep{beyer2022knowledge}. In 
\cref{fig:scaling_iters} we show our reinforced training also benefits from 
longer training, achieving 71.7\% \imagenet{}-val zero-shot accuracy after 120 epochs 
using only a 12M subset of \datacompOneB{}. In comparison, non-reinforced 
training at best reaches {55.7\%}  %
accuracy. 

We also demonstrate scaling with dataset size in \cref{fig:scaling_data}, where 
we deploy subsets of \datacompOneB{} from 1.28M to all 1.28B samples.
For all experiments we train for 20k iterations with global batch size of 65k  (equivalent to one 
epoch training on 1.28B subset). Training on \ourdataset{} reaches above 55.2\% accuracy with 1.28M samples while 
training on \datacompOneB{} gets only to $\sim$6\% accuracy. In this setup, we observe more than 
100$\times$ data efficiency using \ourdataset{}.
Moreover, we observe 1000$\times$ data efficiency for performance on \flickrval{}.

\begin{table*}[t!]
    \centering
    \resizebox{0.99\textwidth}{!}{
        \begin{tabular}{l|cC{1.5cm}|C{2cm}C{1.5cm}C{2cm}C{2cm}|cccccc|C{1.4cm}}
            \toprule[1.5pt]
            \multirow{2}{*}{\textbf{Name}} 
            & \multirow{2}{*}{\textbf{Dataset}}
            & \multirow{2}{1.7cm}{\centering\textbf{Seen Samples}}
            & \multirow{2}{2cm}{\centering\textbf{Image Encoder}}
            & \multirow{2}{1.5cm}{\centering\textbf{Text Encoder}}
            & \multirow{2}{2cm}{\centering\textbf{Params (M) {(img+txt)}}}
            & \multirow{2}{2cm}{\centering\textbf{Latency (ms) {(img+txt)}}}
            & \multicolumn{2}{c}{\textbf{Zero-shot CLS}}
            & \multicolumn{2}{c}{\textbf{\flickrval{} Ret.}}
            & \multicolumn{2}{c}{\textbf{COCO Ret.}}
            & \multirow{2}{1.4cm}{\centering\textbf{Avg. Perf. on 38}}
            \\
            \cmidrule(lr){8-9}
            \cmidrule(lr){10-11}
            \cmidrule(lr){12-13}
             & & & & &&&
             \imagenetval{} & IN-shift &T$\to$I&I$\to$T&T$\to$I&I$\to$T& \\
             \midrule[1.25pt]
                Ensemble Teacher
               & \begin{tabular}{c}
                \datacompOneB{}~\citep{DataComp}\\
                OpenAI-400M~\citep{CLIP}
               \end{tabular}
               & - & 
               \begin{tabular}{c}
                ViT-L/14\\
                ViT-L/14
               \end{tabular}
               &
               \begin{tabular}{c}
                Base\\
                Base
               \end{tabular}
               & (-) & (-) & 
             80.1 & 69.6 & 74.5& 92.3 &46.7 &66.5&67.3\\
             \midrule[1.25pt]

             TinyCLIP-RN19M~\citep{tinyclip} & LAION-400M~\citep{Laion400} &
             15.2B & ResNet-19M & Custom & 18.6 + 44.8 & 1.9 + 1.9 & 56.3 & 43.6 & 58.0 & 75.4 & 30.9 & 47.8 & 48.3 \\

             TinyCLIP-RN30M~\citep{tinyclip} & LAION-400M~\citep{Laion400} &
             15.2B & ResNet-30M & Custom & 29.6 + 54.2 & 2.6 + 2.6 & 59.1 & 45.7 & 61.5 & 80.1 & 33.8 & 51.6 & 50.2 \\

             TinyCLIP-40M/32~\citep{tinyclip} & LAION-400M~\citep{Laion400} &
             15.2B & ViT-40M/32 & Custom & 39.7 + 44.5 & 3.0 + 1.9 & 59.8 & 46.5 & 59.1 & 76.1 & 33.5 & 48.7 & 51.2 \\

             \textbf{\ourmodelSZero{}} & \ourdatasetOneB{} &  13B & \ourimagezero{}&\ourtextzero{} & 11.4 + 42.4 & 1.5 + 1.6 &
             \textbf{67.8} & \textbf{55.1} & \textbf{67.7} & \textbf{85.9} & \textbf{40.4} & \textbf{58.7} & \textbf{58.1}\\

             \midrule
             
             OpenAI-RN50 & OpenAI-400M~\citep{CLIP} &  13B & 
             ResNet-50 & Base 
             & 38.3 + 63.4 & 3.3 + 3.3 & 
             59.8 & 45.1 & 57.4 & 80.0 & 28.5 & 48.8 & 48.1 \\

             TinyCLIP-61M/32~\citep{tinyclip} & LAION-400M~\citep{Laion400} &
             15.2B & ViT-61M/32 & Custom & 61.4 + 54.0 & 4.3 + 2.6 & 62.4 & 48.7 & 62.6 & 78.7 & 36.5 & 52.8 & 53.0 \\

             TinyCLIP-63M/32~\citep{tinyclip}&
             \begin{tabular}{c}
                LAION-400M~\citep{Laion400}\\
                YFCC-15M~\citep{yfcc}
             \end{tabular}
             &  15.8B & ViT-63M/32 & Custom & (-) & (-) & 
             64.5 & (-) & 66.0 & 84.9 & 38.5 & 56.9 & (-) \\

             \textbf{\ourmodelSOne{}} & \ourdatasetOneB{} &  13B &\ourimageone{}& Base & 21.5 + 63.4 & 2.5 + 3.3 &
             \textbf{72.6} & \textbf{60.7} & \textbf{71.0} & \textbf{89.2} & \textbf{44.0} & \textbf{62.2} & \textbf{61.3} \\

             \midrule

             OpenAI-RN101 & OpenAI-400M~\citep{CLIP} &  13B & 
             ResNet-101 & Base 
             & 56.3 + 63.4 & 4.3 + 3.3 & 
             62.3 & 48.5 & 58.0 & 79.0 & 30.7 & 49.8 & 50.3 \\

             OpenAI-B/32 & OpenAI-400M~\citep{CLIP} &  13B & 
             \multirow{3}{*}{ViT-B/32} & \multirow{3}{*}{Base} 
             & \multirow{3}{*}{86.2 + 63.4} & \multirow{3}{*}{5.9 + 3.3} & 
             63.3 & 48.5&58.8&78.9&30.4&50.1&52.5\\
             LAION-B/32 & LAION-2B~\citep{Laion} &  32B & & 
             & & & 
             65.7 & 51.9&66.4&84.4&39.1&56.2&54.8\\
             DataComp-B/32 & \datacompOneB{}~\citep{DataComp} &  13B & & 
             & & & 
             69.2 & 55.2&61.1&79.0&37.1&53.5&58.0\\
             
             DataComp-B/32-256 & \datacompOneB{}~\citep{DataComp} &  34B & ViT-B/32-256 
             & Base & 86.2 + 63.4 & 6.2 + 3.3 & 
             72.8 & 58.7&64.9&84.8&39.9&57.9&60.9\\

             \textbf{\ourmodelSTwo{}} &  \ourdatasetOneB{}  &  13B & \ourimagetwo{} & Base & 35.7 + 63.4 & 3.6 + 3.3 &
             \textbf{74.4} & \textbf{63.1} & \textbf{73.4} & \textbf{90.3} & \textbf{45.4} & \textbf{63.4} & \textbf{63.7} \\

             \midrule
             VeCLIP-B/16~\citep{VeCLIP} &  WIT-200M& 6.4B &  \multirow{9}{*}{ViT-B/16} 
             & Base & 86.2 + 63.4 & 11.5 + 3.3 & 
             64.6 & (-) & 76.3 & 91.1 & 48.4 & 67.2 & (-) \\
             OpenAI-B/16 &  WIT-400M~\citep{CLIP} &  13B &
             & Base & 86.2 + 63.4 & 11.5 + 3.3 & 
             68.3 & 55.9 & 67.7 & 85.9 & 40.4 & 58.7 & 58.1 \\
             LAION-B/16 &  LAION-2B~\citep{Laion} &  34B &  & Base & 86.2 + 63.4 & 11.5 + 3.3 & 
             70.2 & 56.6 & 69.8 & 86.3 & 42.3 & 59.4 & 58.7\\
             EVA02-B/16 & Merged-2B~\citep{EVA-CLIP} &  8B & & Base & 86.2 + 63.4 & (-) & 
             74.7 & 59.6 & 71.5 & 86.0 & 42.2 & 58.7 & 58.9\\
             DFN-B/16 &  DFN-2B~\citep{DFN} &  13B &   & Base & 86.2 + 63.4 & 11.5 + 3.3 & 
             76.2 & 62.3 & 69.1 & 85.4& 43.4 & 60.4& 60.9\\
             DataComp-B/16 & \datacompOneB{}~\citep{DataComp} &  13B &  & Base & 86.2 + 63.4 
             & 11.5 + 3.3 & 
             73.5 & 60.8 & 69.8 & 86.3& 42.3 & 59.4& 61.5\\
             
             SigLIP-B/16~\citep{SigLIP} & Webli-1B &  40B &   & Custom 
             & 92.9 + 110.3 & 9.9 + 5.8 &
             76.0 & 61.0 & 74.7 & 89.1& 47.8 & 65.7& 62.3\\

             \textbf{\ourmodelB{}} & \ourdatasetOneB{} &  13B &   & Base & 86.3 + 63.4 & 10.4 + 3.3 &
             76.8 & 65.6 & \textbf{77.3} & 91.4& \textbf{50.6} & \textbf{68.8} & 65.2\\
             \textbf{\ourmodelB{} (LT)} & \ourdatasetOneB{} &  39B &   & Base & 86.3 
             + 63.4 & 10.4 + 3.3 &
             \textbf{77.2} & \textbf{66.1} & 76.9 & \textbf{92.3} & 50.0 & \textbf{68.7} & \textbf{65.8}\\

            \bottomrule[1.5pt]
        \end{tabular}
    }
    \vspace{-5pt}
    \caption{\textbf{\ourmodel{} family of models has the best
    average performance at various latencies.}
    Retrieval performances are reported @1. 
    Last column shows average performance on 38 datasets as in OpenCLIP~\citep{OpenCLIP}. Models are grouped by their total latency in increasing order and by performance within each group. ``Base'' refers to standard CLIP Transformer-based~\citep{transformers} text encoder with 12 layers, and ``Custom'' stands for customized text encoder used in the respective method. For TinyCLIP-63M/32 and EVA02-B/16, we were unable to reliably benchmark models. \textit{Note}: EVA02-B/16~\citep{EVA-CLIP} uses MIM pretrained weights for its vision encoder and OpenCLIP-B pretrained weights for its text encoder. TinyCLIP models use advanced weight initialization methods utilizing OpenCLIP models trained on LAION-2B\citep{Laion} dataset. All other models, including ours are trained from scratch. ``(LT)'' refers to longer training schedule, described in detail in~\cref{sec:long_train}.
    }
    \label{tab:full_eval}
    \vspace{-5pt}
\end{table*}

\subsection{Comparison with State-of-the-art}\label{sec:comparison}

In \cref{tab:full_eval}, we compare with methods with large scale training. \ourmodelSZero{}, trained on \ourdatasetOneB{} significantly outperforms recent works like TinyCLIP~\citep{tinyclip}, and has similar performance as a ViT-B/32 model trained on DataComp~\citep{DataComp} while being 2.8$\times$ smaller and 3$\times$ faster.
\ourmodelSTwo{} obtains 2.8\% better average performance on 38 datasets and significantly better retrieval performance when compared to ViT-B/32-256 model trained 2.6$\times$ longer on DataComp~\citep{DataComp}. \ourmodelSTwo{} is 1.5$\times$ smaller and 1.4$\times$ faster than ViT-B/32-256 model. \ourmodelB{} obtains 2.9\% better average performance on 38 datasets and better retrieval performance while being 26.3\% smaller than SigLIP-B/16~\citep{SigLIP} model, which is trained approximately 3$\times$ longer on WebLI dataset.

\subsection{Retrieval Performance Analysis}\label{sec:retrieval}
We evaluate our models on the recent Attribute, Relation and Order (ARO) benchmark~\citep{vlmbow}. 
We compare our \ourmodelB{} trained on \ourdatasetOneB{} with all the publicly available ViT-B/16:Base models in \cref{tab:aro_eval}.
Optimizing solely for zero-shot classification or retrieval using noisy webscale datasets can degrade the compositional understanding of natural scenes.
\ourdataset{} largely improves the models performance on ARO benchmark while obtaining good performance on zero-shot classification and retrieval tasks. Compared to the recent SigLIP method~\citep{SigLIP}, \ourmodelB{} obtains 19.5\% and 12.4\% better accuracy on Visual Genome Relation and Attributes datasets and achieves improved recall@1 on Flickr30k-Order and COCO-Order datasets by 69.7\% and 50.3\%, respectively.

\begin{table}[t!]
    \centering
    \resizebox{0.95\linewidth}{!}{
        \begin{tabular}{c@{\hspace{0.95\tabcolsep}}c@{\hspace{0.95\tabcolsep}}|c@{\hspace{0.95\tabcolsep}}c@{\hspace{0.95\tabcolsep}}c@{\hspace{0.95\tabcolsep}}c@{\hspace{0.95\tabcolsep}}c@{\hspace{0.95\tabcolsep}}}
            \toprule[1.5pt]
            \multirow{2}{*}{\textbf{Method}} 
            & \multirow{2}{*}{\textbf{Dataset}}
            & \textbf{\imagenetval{}}
            & \textbf{VG} & \textbf{VG} 
            & \textbf{COCO} & \textbf{\flickrval{}}
            \\
             & & \textbf{zero-shot} & \textbf{Rel.} & \textbf{Attr.} & \textbf{Order} & \textbf{Order} \\
            
             \midrule[1.25pt]

             CLIP & OpenAI-400M~\citep{CLIP} &  
             68.3 & \textbf{58.7} & 62.2 & \underline{50.4} & \underline{57.3} \\

             CLIP & LAION-2B~\citep{Laion} &  
             70.2 & 39.7 & \underline{62.3} & 31.0 & 37.5\\
             
             CLIP & \datacompOneB{}~\citep{DataComp} & 
             73.5 & 35.9 & 57.0 & 29.6 & 35.2 \\
             
             SigLIP~\citep{SigLIP} & Webli-1B &
             76.0 & 35.1 & 56.0 & 32.7 & 40.7 \\
             
             CLIP & DFN-2B~\citep{DFN} &
             \underline{76.2} & 33.1 & 57.4 & 18.5 & 22.5 \\
             
             \textbf{\ourmodelB{}} &  \ourdatasetOneB{} & \textbf{76.8} 
             & \underline{54.6} & \textbf{68.4} & \textbf{55.5} & \textbf{61.2} 
             \\
            
            \bottomrule[1.5pt]
        \end{tabular}
    }
    \vspace{-5pt}
    \caption{\textbf{Performance on ARO benchmark.} All the models use ViT-B/16 as image encoder and the Base text encoder. For VG Rel. and VG Attr. datasets, Macro Acc. is reported and for Flickr30k-Order and COCO-Order recall@1 is reported following~\citep{vlmbow}. 
    }
    \label{tab:aro_eval}
    \vspace{-5pt}
\end{table}

\section{Conclusion}

In this work we introduced \ourmodel{} aligned image-text backbones, designed for on-device CLIP inference (low latency and size). We also introduced \ourdataset{}, a reinforcement of \datacomp{} with knowledge from a pre-trained image captioning model and an ensemble of strong CLIP models. We demonstrated 10$\times$-1000$\times$ learning efficiency with our reinforced dataset. \ourmodel{} models trained on \ourdataset{} obtain state-of-the-art latency-accuracy tradeoff when compared to previous works. \ourmodel{} models also exhibit better robustness and improved performance on Attribute, Relation and Order (ARO) benchmark.

\section*{Acknowledgments}

We thank Jason Ramapuram, Vaishaal Shankar, and Russ Webb for their valuable feedback and discussions. We also thank Albin Madappally Jose for support in dataset creation and processing. Finally, we would like to thank the entire Machine Learning Research team at Apple for helpful discussions and assistance with the infrastructure.

{
    \small
    \bibliographystyle{ieeenat_fullname}
    \bibliography{main}
}

\appendix

\clearpage
\maketitlesupplementary

\section{Image Encoder Configurations}\label{sec:mci_configs}
In our work, we introduce 3 stage configurations for FastViT architecture that substantially improves the model with limited impact on latency. The three configurations are described in~\cref{tab:brief_config}. Comparison of our image encoders with FastViT image encoder when trained on ImageNet-1k dataset in a supervised setting (described in~\cref{sec:supp_exp_setup}) is shown in~\cref{fig:acc_vs_latency_in1k_sup}. 

\begin{table}[h!]
\centering
\begin{subtable}{0.7\columnwidth}
    
  \centering
    \resizebox{0.99\columnwidth}{!}{
  \begin{tabular}{cll}
      \toprule[1.5pt]
    Variant & $\{C_1, C_2, C_3, C_4\}$ & $\{L_1, L_2, L_3, L_4\}$  \\ \midrule
    \ourimagezero{} & $\{64, 128, 256, 512\}$ & $\{2, 6, 10, 2\}$  \\ 
    \ourimageone{} & $\{64, 128, 256, 512\}$ & $\{4, 12, 20, 4\}$ \\ 
    \ourimagetwo{} & $\{80, 160, 320, 640\}$ & $\{4, 12, 24, 4\}$ \\ 
      \bottomrule[1.5pt]
  \end{tabular}}

\end{subtable}
\caption{Configurations of \ourimage{}.}\label{tab:brief_config}
\vspace{-7pt}
\end{table}

\section{Experimental Setup}\label{sec:supp_exp_setup}
Additional details of our training and evaluation are provided in this section. 
\Cref{tab:hyperparams_datacomp1b} summarizes the hyperparameters we used to train \ourmodelB{} on \ourdatasetOneB{}. For other variants of MobileCLIP (S0, S1, and S2) we use the same hyperparameters except using $\lambda=1.0$. For experiments on \ourdatasetTM{} we use global batch size of 8192. All models trained on DataComp(-DR) use strong image augmentation unless stated otherwise.

For our ensemble distillation ablations in \cref{sec:clip_ensembles}, we use 32 
total A100 GPUs but we use the same global batch size of 8192 as our other 
ablations. We also use a smaller uniformly sampled DataComp-8M for ablations in \cref{sec:clip_ensembles,sec:augmentation_ablation} that results in 
a slightly lower performance than \ourdatasetTM{} used for the rest of 
ablations.

The seen samples reported for \ourdataset{} is a triplet of one randomly augmented image, one ground-truth caption, and one randomly picked synthetic caption. The reported number of iterations is the number of seen samples divided by the global batch size.

\begin{figure}
    \centering
    \includegraphics[width=0.85\linewidth]{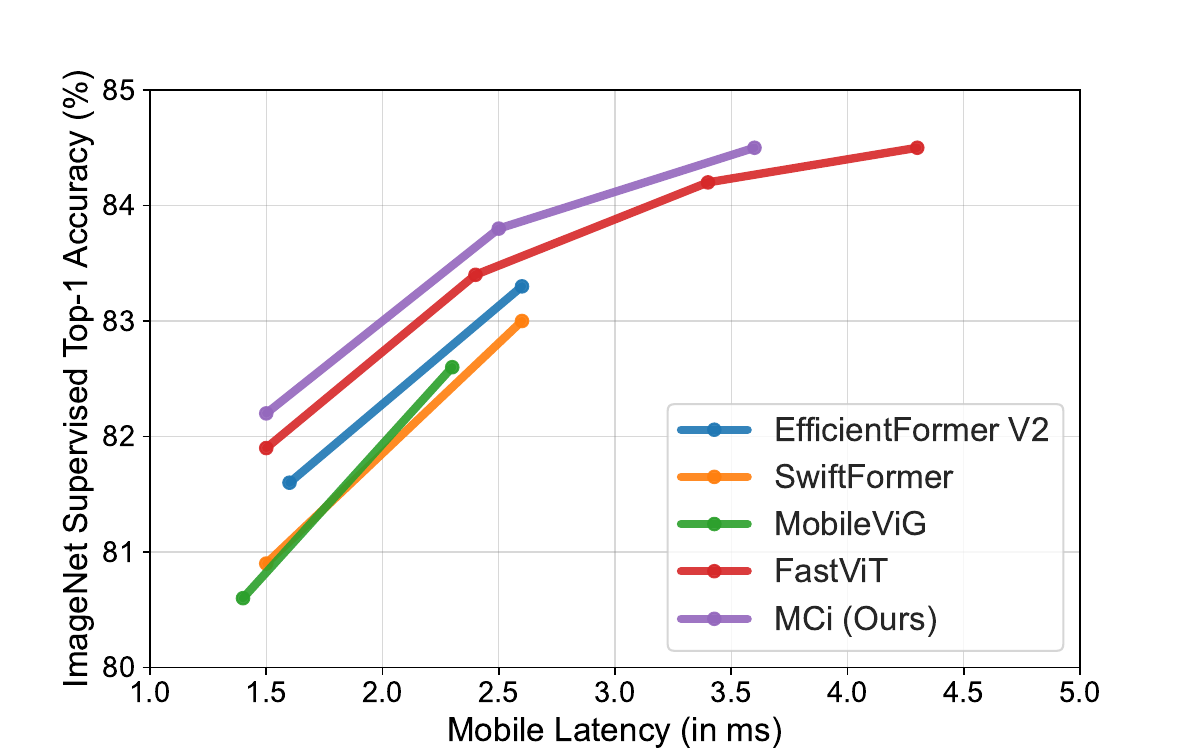}
    \caption{
    Top-1 Accuracy on \imagenet{} v/s latency plot of \ourimage{} variants and recent state-of-the-art architectures.}
    \label{fig:acc_vs_latency_in1k_sup}
\vspace{-5pt}
\end{figure}

For ImageNet-1k experiments, we follow the training recipe prescribed in~\citep{Deit, efficientformerv2}, i.e. the models are trained for 300 epochs using AdamW optimizer with weight decay of 0.05 and peak learning rate $10^{-3}$ for a total batch size of 1024. The number of warmup epochs is set to 5 and cosine schedule is used to decay the learning rate. The teacher model for distillation is RegNetY-16GF~\citep{RegNet} Our implementation uses Timm library~\citep{timm} and all the models were trained on single machine with 8$\times$NVIDIA A100 GPUs. The hyperparameters for the three variants of \ourimage{} are detailed in \cref{tab:hyperparams_inet1k}. The performance of \ourimage{} variants is detailed in \cref{tab:ImageNet_1k} and compared against recent state-of-art efficient architectures. \ourimage{} obtains the best trade-off amongst recent efficient architectures as seen in \cref{fig:acc_vs_latency_in1k_sup}.

\begin{table}[h]
\centering
\scalebox{0.9}{
\begin{tabular}{l|c}
\toprule
\multirow{2}{*}{Hyperparameter}   & Training \\
 & \ourimagezero{}, \ourimageone{}, \ourimagetwo{} \\
\midrule
Stochastic depth rate & [0.0, 0.05, 0.15]  \\
Input resolution & 256$\times$256  \\
Data augmentation & RandAugment  \\
Mixup $\alpha$ & 0.8 \\
CutMix $\alpha$ & 1.0 \\
Random erase prob. & 0.25 \\
Label smoothing & 0.1 \\
Train epochs & 300 \\
Warmup epochs & 5 \\
Batch size & 1024 \\
Optimizer & AdamW \\
Peak learning rate & 1e-3 \\
LR. decay schedule & cosine \\
Weight decay rate & 0.05 \\
Gradient clipping & \xmark \\
EMA decay rate & 0.9995 \\

\bottomrule

\end{tabular}
}
\caption{Training hyperparameters for ImageNet-1k experiments.}
\label{tab:hyperparams_inet1k}
\end{table}

\begin{table}[]
    \centering

    \scalebox{0.72}{
    \begin{tabular}{l | c c c c | c}
    \toprule
    \multirow{2}{*}{Model}        & Eval    & Param     & FLOPs     & Mobile       & Top-1  \\ 
                                    & Image &        &        & Latency  &  Acc.      \\   
                    & Size & (M) & (G)  & (ms) & (\%) \\
                                    
    \midrule

    MobileViG-M~\citep{mobilevig} & 224 & 14.0 & 1.5 & 1.4 & 80.6\\
    SwiftFormer-L1~\citep{swiftformer} & 224 & 12.1 & 1.6 & 1.5 & 80.9 \\
    EfficientFormerV2-S2~\citep{efficientformerv2} & 224 & 12.6 & 1.3 & 1.6 & 81.6 \\ 
    FastViT-SA12~\citep{vasu2023fastvit} & 256 & 11.5 & 1.9 & 1.5 & 81.9 \\ 
    \textbf{\ourimagezero{} (ours)}       & 256 & 11.8 & 2.4 & 1.5 & 82.2 \\ 
    
    \midrule

    MobileViG-B~\citep{mobilevig} & 224 & 26.7 & 2.8 & 2.3 & 82.6 \\
    SwiftFormer-L3~\citep{swiftformer} & 224 & 28.5 & 4.0 & 2.6 & 83.0 \\
    EfficientFormerV2-L~\citep{efficientformerv2} & 224 & 26.1 & 2.6 & 2.6 & 83.3 \\
    FastViT-SA24~\citep{vasu2023fastvit} & 256 & 21.5 & 3.8 & 2.4 & 83.4 \\ 
    \textbf{\ourimageone{} (ours)}       & 256 & 21.9 & 4.7 & 2.5 & 83.8 \\

    \midrule
    
    FastViT-MA36~\citep{vasu2023fastvit} & 256 & 43.9 & 7.8 & 4.3 & 84.5 \\ 
    \textbf{\ourimagetwo{} (ours)}       & 256 & 36.3 & 7.8 & 3.6 & 84.5 \\
    
    \bottomrule
    \end{tabular}
    
    }
        \caption{Comparison of \ourimage{} variants with recent state-of-the-art models on \imagenet{} classification task.}
    \label{tab:ImageNet_1k}
\end{table}

\begin{table}[h]
\centering
\scalebox{0.85}{
\begin{tabular}{l|c}
\toprule
\multirow{2}{*}{Hyperparameter}   & Value \\
& MobileCLIP-B, S0, S1, S2 \\
\midrule
Input resolution & 224$^2$, 256$^2$, 256$^2$, 256$^2$  \\
Context length & 77\\
Data augmentation & RandAugment  \\
Random resize crop scale & [0.08, 1.0] \\
Random resized crop ratio & [0.75, 1.33] \\
RangeAugment target value & (40, 20) \\
Train iterations & 200k \\
Warmup iterations & 2k \\
Global batch size & 65536 \\
Optimizer & AdamW \\
AdamW beta1 & 0.9\\
AdamW beta2 & 0.95\\
Max learning rate & 1e-3 \\
Min learning rate & 1e-6 \\
LR. decay schedule & cosine \\
Weight decay rate & 0.2 \\
Gradient clipping & \xmark \\
Mixed precision& BFloat16 \\
EMA decay rate & 0.9995 \\
CLIP loss weight & 0.25\\
KD loss weight & 0.75\\
GT caption weight & 1.0\\
Synth.\ caption weight & 1.0\\
Synth.\ teacher & coca\_ViT-L-14\\
Teacher 1 & openai-ViT-L-14 \\
Teacher 2 & datacomp\_xl\_s13b\_b90k-ViT-L-14 \\
Teacher resolution & 224$\times$224\\
\bottomrule

\end{tabular}
}
\caption{Training hyperparameters for our CLIP experiments on \ourdataset{}.}
\label{tab:hyperparams_datacomp1b}
\end{table}

\section{Image Augmentation}
\label{sec:augmentation_ablation}
In this section we provide a detailed ablation on the effect of image augmentations. The training setup is the same as training with \ourdatasetTM{} presented in \cref{sec:dr12m}, except we used an 8M subset for this ablation.  In \cref{tab:teacher_aug_ablation} we show classification and retrieval performance of a ViT-B/16 based CLIP model trained with our final loss as in \cref{eq:final_total_loss} ($\lambda=1$) and different image augmentations. Note that we feed the same augmented image to both teacher and student models. First, we consider \verb|RandomResizedCrop| (RRC) with three magnitudes (0.08, 0.4, 0.9) determining the lower bound of random area of the crop (smaller lower bound means stronger augmentation). We observe that strong RRC results in significant accuracy improvement both for classification and retrieval metrics. While using strong RRC augmentation is standard for supervised training, for CLIP training the widely used recipe~\citep{CLIP} includes weak RRC (lower-bound for scale= 0.9).

We further utilize \verb|RangeAugment|~\citep{RangeAug} to automatically adjust Brightness, Contrast, and Noise. We use PSNR metric with target range [20, 40] and a Cosine curriculum. Since in \verb|RangeAugment| individual augmentation magnitudes are adjusted dynamically during training, they cannot be stored as part of the dataset reinforcement process. Hence, we only apply it to images fed to the student model. We show that if the same augmentation is applied to both student and teacher (not feasible for our dataset reinforcement approach) further improvement can be obtained (56.6\% vs 55.9\% on \imagenet{}-val).

Finally, we consider \verb|RandomHorizontalFlip|, \verb|RandomErasing|~\citep{RandErase}, and  \verb|RandAugment|~\citep{RandAug}, and find that only \verb|RandAugment| is beneficial in our setup. Our reinforced datasets include parameters of RRC and \verb|RandAugment| and during training time we apply \verb|RangeAugment| to images fed to the student model.

\begin{table}[h]
    \centering
    \resizebox{0.5\textwidth}{!}{
        \begin{tabular}{c|cccccc|c}
            \toprule[1.5pt]
            \multirow{2}{*}{\textbf{Image Augmentations}} &
            \multicolumn{2}{c}{\textbf{Zero-shot CLS}} &
            \multicolumn{2}{c}{\textbf{Flickr30k Ret.}} &
            \multicolumn{2}{c}{\textbf{COCO Ret.}} &
            \textbf{Avg Perf.}
            \\
            \cmidrule(lr){2-3}
            \cmidrule(lr){4-5}
            \cmidrule(lr){6-7}
             \begin{tabular}{c}
                  \\
             \end{tabular}
             &
             \imagenetval{} & IN-shift &
             I2T&T2I&
             I2T&T2I& 
             \textbf{on 38}\\
             \midrule[1.25pt]
             
             \rowcolor{lightgray}
             \begin{tabular}{c}
                  RandomResizedCrop: \texttt{0.9-1.0}\\
                  Student-RangeAugment~\citep{RangeAug}
             \end{tabular}
             &
             51.0 & 40.1 & 54.2& 68.5 &30.5 &45.3 &45.9 \\

             \begin{tabular}{c}
                  RandomResizedCrop: \texttt{0.4-1.0}\\
                  Student-RangeAugment
             \end{tabular}
             &
             55.0 & 43.9 & 60.4& 76.0 &34.1 &48.4 &48.9 \\

            \rowcolor{lightgray}
             \begin{tabular}{c}
                  RandomResizedCrop: \texttt{0.08-1.0}\\
                  Student-RangeAugment
             \end{tabular}
             &
             55.9 & 44.6 & 58.8& 76.1 &34.2 &49.0 &49.6 \\

            \midrule
            
             \begin{tabular}{c}
                  RandomResizedCrop: \texttt{0.08-1.0}\\
             \end{tabular}
             &
             56.4 & 44.6 & 59.8& 74.6 &34.4 &49.3 &49.1 \\

            \rowcolor{lightgray}
             \begin{tabular}{c}
                  RandomResizedCrop: \texttt{0.08-1.0}\\
                  Student\&Teacher-RangeAugment
             \end{tabular}
             &
             56.6 & 44.9 & 60.2& 74.0 &34.9 &50.5 &50.8 \\

            \midrule
            \begin{tabular}{c}
                  RandomResizedCrop: \texttt{0.08-1.0}\\
                  Student-RangeAugment\\
                  RandomHorizontalFlip: \texttt{p=0.5}
             \end{tabular}
             &
             55.9 & 44.7 & 59.4& 75.9 &34.4 &49.2 &48.8 \\

            \rowcolor{lightgray}
             \begin{tabular}{c}
                  RandomResizedCrop: \texttt{0.08-1.0}\\
                  Student-RangeAugment\\
                  RandomErasing~\citep{RandErase}: \texttt{p=0.25}
             \end{tabular}
             &
             55.8 & 44.5 & 59.4& 75.3 &34.5 &49.7 &49.1 \\
             \rowcolor{blue!25}
             \begin{tabular}{c}
                  RandomResizedCrop: \texttt{0.08-1.0}\\
                  Student-RangeAugment\\
                  RandAugment~\citep{RandAug}
             \end{tabular}
             &
             56.6 & 45.4 & 60.9& 78.3 &35.0 &51.0 &50.2 \\
             
            \bottomrule[1.5pt]
        \end{tabular}
    }
    \caption{Ablation on different augmentations for distillation.
    We highlight our choice with {\colorbox{blue!25}{blue}}.}
  \label{tab:teacher_aug_ablation}
\end{table}

\section{CLIP Ensembles}
\label{sec:clip_ensembles}
In this section we provide a detailed ablation on CLIP ensembles. First, we show that we can construct more accurate zero-shot models by ensembling pretrained individual CLIP models. For inference, we concatenate normalized embeddings of each modality followed by a  re-normalization. In \cref{tab:teacher_ablation_teacher} we show performance of some CLIP ensemble models that we picked from OpenCLIP~\citep{OpenCLIP}. We also include performance of individual models. Evidently, ensembling results in improved performance. For example, an ensemble of two pretrained \verb|ViT-L-14|-based CLIP models from \verb|datacomp_xl_s13b_b90k| and \verb|openai| results in average performance of 67.3\%, while each individual model has 66.3\% and 61.7\% performance, respectively. Further, ensembling can be a more parameter efficient approach to obtain a stronger model. For instance, the ensemble of two \verb|ViT-L-14|-based CLIP models has less parameters than the one with \verb|ViT-bigG-14| image encoder, but comes with the same \imagenet{}-val performance (80.1\%). In general, given a set of pretrained CLIP models (e.g., as in OpenCLIP~\citep{OpenCLIP}) with this approach we can push state-of-the-art and obtain stronger zero-shot performance. Here, we show and ensemble of four CLIP models can reach up to 81.7\% zero-shot classification performance on \imagenet{}-val, while individual models' performance is not more than 80.1\%. As stronger individual models become publicly available, one can create stronger ensembles with this approach.

In this work, we are interested in creating a strong ensemble model to be used as a teacher in the context of distillation. In \cref{tab:teacher_ablation_student} we show performance of a ViT-B/16 CLIP model trained with different CLIP models as teacher (both individual models and ensembles). Training setup is the same as that of in \cref{sec:dr12m}, except we use a uniformly sampled 8M subset. Similar to standard distillation for classification task~\citep{hinton2015distilling}, we observe that more accurate CLIP models are not necessarily better teachers. We picked the ensemble of two \verb|ViT-L-14|-based CLIP models as the teacher model (highlighted in blue) in our dataset reinforcement process.

\begin{table*}[h]
    \centering
    \resizebox{0.8\textwidth}{!}{
        \begin{tabular}{ccc|cccccc|c}
            \toprule[1.5pt]
            \textbf{Teacher} &
            \textbf{Teacher} &
            \textbf{Teacher} &
            \multicolumn{2}{c}{\textbf{Zero-shot CLS}} &
            \multicolumn{2}{c}{\textbf{Flickr30k Ret.}} &
            \multicolumn{2}{c}{\textbf{COCO Ret.}} &
            \textbf{Avg Perf.}
            \\
            \cmidrule(lr){4-5}
            \cmidrule(lr){6-7}
            \cmidrule(lr){8-9}
             \textbf{Models(s)}&
             \textbf{Pre-taining(s)}&
             \textbf{Resolution(s)}& 
             \imagenetval{} & IN-shift &
             I2T&T2I&
             I2T&T2I& 
             \textbf{on 38}\\
             \midrule[1.25pt]
             \rowcolor{lightgray}
             \texttt{ViT-bigG-14} & \texttt{laion2b\_s39b\_b160k} & 224&
             80.1 & 69.1 & 79.6& 92.9 &51.4 &67.4&66.7 \\
             \texttt{EVA01-g-14-plus} & \texttt{merged2b\_s11b\_b114k} & 224&
             79.3 & 69.3 & 79.0& 91.7 &50.3 &68.2&66.2 \\
             \rowcolor{lightgray}
             \texttt{ViT-L-14} & \texttt{datacomp\_xl\_s13b\_b90k} & 224&
             79.2 & 67.9 & 73.4& 89.0&45.7 &63.3 &66.3 \\
             \texttt{ViT-L-14} & \texttt{openai} & 224&
             75.5 & 64.9 & 65.0& 85.2 &36.5 &56.3&61.7 \\
             \rowcolor{lightgray}
             \texttt{ViT-L-14-336} & \texttt{openai} & 336&
             76.6 & 67.1 & 66.9& 87.7 &37.1 &57.9&62.8 \\
             
             \midrule[1.25pt]
             \rowcolor{blue!25}
             \begin{tabular}{c}\texttt{ViT-L-14}\\
             \texttt{ViT-L-14} \end{tabular}& 
             \begin{tabular}{c}\texttt{datacomp\_xl\_s13b\_b90k}\\
             \texttt{openai} \end{tabular} 
             & \begin{tabular}{c}224\\224\end{tabular} &
             80.1 & 69.6 & 74.5& 92.3 &46.7 &66.5&67.3 \\
             
             \rowcolor{lightgray}
             \begin{tabular}{c}\texttt{ViT-L-14}\\
             \texttt{ViT-L-14-336} \end{tabular}& 
             \begin{tabular}{c}\texttt{datacomp\_xl\_s13b\_b90k}\\
             \texttt{openai} \end{tabular} 
             & \begin{tabular}{c}224\\336\end{tabular} &
             80.5 & 70.6 & 75.8& 91.8 &47.0 &67.0&67.8 \\
             
             \midrule[1.25pt]
             
             \begin{tabular}{c}
             \texttt{EVA01-g-14-plus}\\
             \texttt{ViT-L-14}\\
             \texttt{ViT-L-14} \end{tabular}& 
             \begin{tabular}{c}
             \texttt{merged2b\_s11b\_b114k}\\
             \texttt{datacomp\_xl\_s13b\_b90k}\\
             \texttt{openai} \end{tabular} 
             & \begin{tabular}{c}
             224\\
             224\\
             224
             \end{tabular} &
             81.1 & 70.9 & 78.1& 93.8 &50.2 &69.7&68.5 \\

            \rowcolor{lightgray}
             \begin{tabular}{c}
             \texttt{EVA01-g-14-plus}\\
             \texttt{ViT-L-14}\\
             \texttt{ViT-L-14-336} \end{tabular}& 
             \begin{tabular}{c}
             \texttt{merged2b\_s11b\_b114k}\\
             \texttt{datacomp\_xl\_s13b\_b90k}\\
             \texttt{openai} \end{tabular} 
             & \begin{tabular}{c}
             224\\
             224\\
             336
             \end{tabular} &
             81.2 & 71.6 & 78.8& 93.7 &50.2 &69.9&68.9 \\

             \begin{tabular}{c}
             \texttt{convnext\_xxlarge}\\
             \texttt{ViT-L-14}\\
             \texttt{ViT-L-14-336} \end{tabular}& 
             \begin{tabular}{c}
             \texttt{laion2b\_s34b\_b82k\_augreg\_soup}\\
             \texttt{datacomp\_xl\_s13b\_b90k}\\
             \texttt{openai} \end{tabular} 
             & \begin{tabular}{c}
             256\\
             224\\
             336
             \end{tabular} &
             81.5 & 71.7 & 79.0& 94.5 &50.5 &69.5&68.7 \\

             \midrule[1.25pt]

             \rowcolor{lightgray}
             \begin{tabular}{c}
             \texttt{ViT-bigG-14}\\
             \texttt{EVA01-g-14-plus}\\
             \texttt{ViT-L-14}\\
             \texttt{ViT-L-14} \end{tabular}& 
             \begin{tabular}{c}
             \texttt{laion2b\_s39b\_b160k}\\
             \texttt{merged2b\_s11b\_b114k}\\
             \texttt{datacomp\_xl\_s13b\_b90k}\\
             \texttt{openai} \end{tabular} 
             & \begin{tabular}{c}
             224\\
             224\\
             224\\
             224
             \end{tabular} &
             81.6 & 71.7 & 79.9& 94.6 &52.4 &71.3&69.4 \\

             \begin{tabular}{c}
             \texttt{EVA01-g-14-plus}\\
             \texttt{ViT-L-14-336}\\
             \texttt{ViT-L-14}\\
             \texttt{convnext\_xxlarge} \end{tabular}& 
             \begin{tabular}{c}
             \texttt{merged2b\_s11b\_b114k}\\
             \texttt{openai}\\
             \texttt{datacomp\_xl\_s13b\_b90k}\\
             \texttt{laion2b\_s34b\_b82k\_augreg\_soup} \end{tabular} 
             & \begin{tabular}{c}
             224\\
             336\\
             224\\
             256
             \end{tabular} &
             81.7 & 72.1 & 80.0& 95.0 &52.0 &70.8&69.3 \\

            \rowcolor{lightgray}
             \begin{tabular}{c}
             \texttt{ViT-L-14}\\
             \texttt{ViT-L-14-336}\\
             \texttt{RN50x64}\\
             \texttt{RN50x16}\\
             \end{tabular}& 
             \begin{tabular}{c}
             \texttt{openai}\\
             \texttt{openai}\\
             \texttt{openai}\\
             \texttt{openai} \end{tabular} 
             & \begin{tabular}{c}
             224\\
             336\\
             384\\
             448
             \end{tabular} &
             78.2 & 68.9 & 73.4& 89.7 &42.0 &63.5&65.5 \\
             
            \bottomrule[1.5pt]
        \end{tabular}
    }
    \caption{Zero-shot evaluation of (ensemble of) clip models. Each group of rows corresponds to an ensemble teacher. All models are taken from OpenCLIP~\citep{OpenCLIP} on \texttt{Aug-2023}.
  We highlight our choice with {\colorbox{blue!25}{blue}}.}
  \label{tab:teacher_ablation_teacher}
\end{table*}

\begin{table*}[t!]
    \centering
    \resizebox{0.99\textwidth}{!}{
        \begin{tabular}{ccc|cccccc|c}
            \toprule[1.5pt]
            \textbf{Teacher} &
            \textbf{Teacher} &
            \textbf{Teacher} &
            \multicolumn{2}{c}{\textbf{Zero-shot CLS}} &
            \multicolumn{2}{c}{\textbf{Flickr30k Ret.}} &
            \multicolumn{2}{c}{\textbf{COCO Ret.}} &
            \textbf{Avg Perf.}
            \\
            \cmidrule(lr){4-5}
            \cmidrule(lr){6-7}
            \cmidrule(lr){8-9}
             \textbf{Models(s)}&
             \textbf{Pre-taining(s)}&
             \textbf{Resolution(s)}& 
             \imagenetval{} & IN-shift &
             I2T&T2I&
             I2T&T2I& 
             \textbf{on 38}\\
             \midrule[1.25pt]
             \rowcolor{lightgray}
             \texttt{ViT-bigG-14} & \texttt{laion2b\_s39b\_b160k} & 224&
             53.4 & 42.6 & 59.6& 76.2 &35.8 &52.1 &47.8 \\
             \texttt{EVA01-g-14-plus} & \texttt{merged2b\_s11b\_b114k} & 224&
             54.5 & 43.3 & 59.6& 74.6 &35.4 &50.8 &47.7 \\
             \rowcolor{lightgray}
             \texttt{ViT-L-14} & \texttt{datacomp\_xl\_s13b\_b90k} & 224&
             54.0 & 43.4 & 58.9& 74.3 &34.3 &50.1 &48.3 \\
             \texttt{ViT-L-14} & \texttt{openai} & 224&
             54.4 & 42.7 & 54.5& 69.1 &29.7 &44.6 &47.2 \\
             \rowcolor{lightgray}
             \texttt{ViT-L-14-336} & \texttt{openai} & 336&
             54.2 & 43.3 & 53.6& 68.7 &30.1 &44.3 &47.2 \\
             
             \midrule[1.25pt]
             
             \rowcolor{blue!25}
             \begin{tabular}{c}\texttt{ViT-L-14}\\
             \texttt{ViT-L-14} \end{tabular}& 
             \begin{tabular}{c}\texttt{datacomp\_xl\_s13b\_b90k}\\
             \texttt{openai} \end{tabular} 
             & \begin{tabular}{c}224\\224\end{tabular} &
             56.3 & 44.8 & 59.2& 74.5 &34.4 &49.9 &49.6 \\
             
             \rowcolor{lightgray}
             \begin{tabular}{c}\texttt{ViT-L-14}\\
             \texttt{ViT-L-14-336} \end{tabular}& 
             \begin{tabular}{c}\texttt{datacomp\_xl\_s13b\_b90k}\\
             \texttt{openai} \end{tabular} 
             & \begin{tabular}{c}224\\336\end{tabular} &
             55.9 & 44.6 & 58.8& 76.1 &34.2 &49.0&49.6 \\
             
             \midrule[1.25pt]
             
             \begin{tabular}{c}
             \texttt{EVA01-g-14-plus}\\
             \texttt{ViT-L-14}\\
             \texttt{ViT-L-14} \end{tabular}& 
             \begin{tabular}{c}
             \texttt{merged2b\_s11b\_b114k}\\
             \texttt{datacomp\_xl\_s13b\_b90k}\\
             \texttt{openai} \end{tabular} 
             & \begin{tabular}{c}
             224\\
             224\\
             224
             \end{tabular} &
             56.2 & 45.0 & 59.6& 76.9 &35.7 &51.5 &49.4 \\

            \rowcolor{lightgray}
             \begin{tabular}{c}
             \texttt{EVA01-g-14-plus}\\
             \texttt{ViT-L-14}\\
             \texttt{ViT-L-14-336} \end{tabular}& 
             \begin{tabular}{c}
             \texttt{merged2b\_s11b\_b114k}\\
             \texttt{datacomp\_xl\_s13b\_b90k}\\
             \texttt{openai} \end{tabular} 
             & \begin{tabular}{c}
             224\\
             224\\
             336
             \end{tabular} &
             56.0 & 44.5 & 60.1& 76.5 &35.3 &50.6 &49.5 \\

             \begin{tabular}{c}
             \texttt{convnext\_xxlarge}\\
             \texttt{ViT-L-14}\\
             \texttt{ViT-L-14-336} \end{tabular}& 
             \begin{tabular}{c}
             \texttt{laion2b\_s34b\_b82k\_augreg\_soup}\\
             \texttt{datacomp\_xl\_s13b\_b90k}\\
             \texttt{openai} \end{tabular} 
             & \begin{tabular}{c}
             256\\
             224\\
             336
             \end{tabular} &
             55.8 & 44.4 & 59.4& 75.1 &35.0 &49.5&50.1 \\

             \midrule[1.25pt]

             \rowcolor{lightgray}
             \begin{tabular}{c}
             \texttt{ViT-bigG-14}\\
             \texttt{EVA01-g-14-plus}\\
             \texttt{ViT-L-14}\\
             \texttt{ViT-L-14} \end{tabular}& 
             \begin{tabular}{c}
             \texttt{laion2b\_s39b\_b160k}\\
             \texttt{merged2b\_s11b\_b114k}\\
             \texttt{datacomp\_xl\_s13b\_b90k}\\
             \texttt{openai} \end{tabular} 
             & \begin{tabular}{c}
             224\\
             224\\
             224\\
             224
             \end{tabular} &
             56.3 & 44.6 & 60.8& 76.2 &35.8 &51.4 &49.2 \\

             \begin{tabular}{c}
             \texttt{EVA01-g-14-plus}\\
             \texttt{ViT-L-14-336}\\
             \texttt{ViT-L-14}\\
             \texttt{convnext\_xxlarge} \end{tabular}& 
             \begin{tabular}{c}
             \texttt{merged2b\_s11b\_b114k}\\
             \texttt{openai}\\
             \texttt{datacomp\_xl\_s13b\_b90k}\\
             \texttt{laion2b\_s34b\_b82k\_augreg\_soup} \end{tabular} 
             & \begin{tabular}{c}
             224\\
             336\\
             224\\
             256
             \end{tabular} &
             55.9 & 44.6 & 60.4& 75.1 &35.6 &52.3&49.4 \\

            \rowcolor{lightgray}
             \begin{tabular}{c}
             \texttt{ViT-L-14}\\
             \texttt{ViT-L-14-336}\\
             \texttt{RN50x64}\\
             \texttt{RN50x16}\\
             \end{tabular}& 
             \begin{tabular}{c}
             \texttt{openai}\\
             \texttt{openai}\\
             \texttt{openai}\\
             \texttt{openai} \end{tabular} 
             & \begin{tabular}{c}
             224\\
             336\\
             384\\
             448
             \end{tabular} &
             56.4 & 44.6 & 57.9& 72.0 &31.7 &47.0&48.6 \\
             
            \bottomrule[1.5pt]
        \end{tabular}
    }
    \caption{Ablation on using different (ensemble of) teacher models in our 
    multi-modal distillation. Each group of rows demonstrate an ensemble 
    teacher. Student architecture is fixed to ViT-B/16 for image encoder and 
    base 12-layer Transformer for text encoder (\ourmodelB{} setup). For this 
    ablation, we use an 8M subset of DataComp and train all experiments for 20k 
    iterations with global batch size of 8k. All models are imported from 
    OpenCLIP~\citep{OpenCLIP} on \texttt{Aug-2023}.
We highlight our choice with {\colorbox{blue!25}{blue}}.
  }
  \label{tab:teacher_ablation_student}
\end{table*}

\section{Ablations on Lossy Compressions}\label{sec:ablations_lossy}

In general, the storage size of datasets depends 
on the file format and the tradeoff between load time and the compression rate. 
In \cref{tab:storage_size} we presented the storage sizes for \ourdatasetTM{} 
and \ourdatasetOneB{} with BFloat16 compression of the embeddings. In this 
section, we further analyze the storage reduction by
i) reducing the number of augmentations, and
ii) lossy compression of embeddings.

We report the total storage size for 12.8k samples of \ourdataset{} in 
\cref{tab:storage_size_full}. The storage size for \ourdatasetTM{} can be 
easily deduced by multiplying the numbers by 1000 (TBs instead of GBs) and by 
$10^5$ for \ourdatasetOneB{}.

\Cref{tab:num_aug_ablation_bfloat16} shows the accuracy of training with 
BFloat16 embeddings achieves accuracies within the standard deviation of the 
training on DataComp-12M.

\begin{table}
    \centering
    \resizebox{0.99\columnwidth}{!}{
        \begin{tabular}{cccC{1cm}C{1cm}C{1cm}C{1.2cm}C{1.2cm}|C{0.7cm}}
            \toprule[1.5pt]
            Image  & Text   & Syn.   & Aug. Params & Text Emb. & Image Emb.  & BFloat16 & Sparsity & Size (GBs)\\
            \midrule[1.25pt]
            \cmark & \cmark & \xmark & \xmark      & \xmark    & \xmark      & \xmark   & \xmark   & 0.9\\
            \midrule
            \cmark & \cmark & \cmark & \cmark      & \xmark    & \xmark      & \xmark   & \xmark   & 0.9\\
            \cmark & \cmark & \cmark & \cmark      & 5+1       & 30          & \xmark   & \xmark   & 3.3\\
            \cmark & \cmark & \cmark & \cmark      & 5+1       & 30          & \cmark   & \xmark   & 1.9\\
            \cmark & \cmark & \cmark & \cmark      & 5+1       & 30          & \xmark   & 50\%     & 1.8\\
            \cmark & \cmark & \cmark & \cmark      & 5+1       & 30          & \cmark   & 50\%     & 1.3\\
            \cmark & \cmark & \cmark & \cmark      & 5+1       & 10          & \xmark   & \xmark   & 1.9\\
            \cmark & \cmark & \cmark & \cmark      & 5+1       & 10          & \cmark   & \xmark   & 1.4\\
            \cmark & \cmark & \cmark & \cmark      & 5         & 5           & \xmark   & \xmark   & 1.5\\
            \cmark & \cmark & \cmark & \cmark      & 5         & 5           & \cmark   & \xmark   & 1.2\\
            \cmark & \cmark & \cmark & \cmark      & 2         & 2           & \xmark   & \xmark   & 1.1\\
            \cmark & \cmark & \cmark & \cmark      & 2         & 2           & \cmark   & \xmark   & 1.0\\
            \bottomrule[1.5pt]
        \end{tabular}
    }
    \caption{Total storage for 12.8k samples stored in individual Pickle Gzip 
    files. Storage for 12.8M and 1.28B samples are approximately the same 
    numbers in TBs and 100 TBs.}
    \label{tab:storage_size_full}
\end{table}

\begin{table}[h]
    \centering
    \resizebox{0.99\columnwidth}{!}{
        \begin{tabular}{c|cccccccc}
            \toprule[1.5pt]
            Num. Aug. &
            1    & 2    & 5    & 10   & 15   & 20   & 25   & 30\\
            \midrule
            w/o BFloat16 &
            60.63 & 63.27 & 64.81 & 64.74 & 64.49 & 64.92 & 64.78 & 64.74\\
            w/ BFloat16 &
             -    &   -   & 64.32 & 64.88 & 64.57 & 64.81 & 65.13 & 64.91\\
            \bottomrule[1.5pt]
        \end{tabular}
    }
    \caption{Effect of BFloat16 and the number of augmentations on \imagenet{}-val zero-shot Accuracy.
We train on \ourdatasetTM{} for approximately 30 epochs. }
    \label{tab:num_aug_ablation_bfloat16}
\end{table}

\section{Hybrid Text Encoder}\label{sec:hybrid_text_encoder}
In this section, we ablate over kernel dimensions for our hybrid text encoder. For this ablation, we use a 6-layered fully convolutional text encoder and systematically increase the kernel size. We use ViT-B/16 as the image encoder for these runs. These models were trained on \ourdatasetTM{} for 30k iterations. From  \cref{tab:text_encoder_kernel_size}, we notice that zero-shot \imagenetval{} performance does improve with increased kernel size, but it is significantly more expensive to run the model on mobile device. For zero-shot \imagenetval{} performance improvement of 1.1\%, the model is 4.5$\times$ slower. From  \cref{tab:text_encoder_kernel_size}, kernel size of 11 obtains the best accuracy-latency trade-off.

For the hybrid design, we use depth-wise 2D convolutional layers. We reshape the 3 dimensional input tensor to \texttt{(BC1S)} format, i.e. \texttt{(Batch Size, Channel Dim., 1, Seq. length)} before feeding the tensor to the convolutional layer. For CLIP, the sequence length is set to 77. The depth-wise convolutions enable interactions between tokens across the sequence. The FFN layers enable interactions between token's channel dimensions. Since the convolution layer is 2D, we simply reuse the reparameterization process described in~\citep{vasu2023fastvit}. 

\begin{table}[h!]
    \centering
    \resizebox{0.6\columnwidth}{!}{
        \begin{tabular}{c|cccccc}    
            \toprule[1.5pt]
            Kernel Size & 3 & \cellcolor{blue!25}11 & 31 \\
            \midrule
            Num Params. (M) & 38.2 & \cellcolor{blue!25}38.3 & 38.4 \\
            Latency (ms) & 1.0 & \cellcolor{blue!25}1.2 & 5.4 \\
            \midrule
            \imagenetval{} & 56.3 & \cellcolor{blue!25}57.9 & 59.0 \\
            \bottomrule[1.5pt]
        \end{tabular}
    }
    \caption{Ablation on kernel size for text encoder.  We train for 30k iterations. We highlight our choice with {\colorbox{blue!25}{blue}}}
    \label{tab:text_encoder_kernel_size}
\end{table}

\begin{table}[h]
    \centering
    \resizebox{1.0\columnwidth}{!}{
        \begin{tabular}{l| c | C{2cm} C{2cm}|C{2cm}|c}
            \toprule[1.5pt]
            \textbf{Image Enc.}
            & \textbf{Dataset}
            & \textbf{\# Image Enc. Params {(M)}}
            & \textbf{Latency (ms) {(img+txt)}}
            & \textbf{0-shot \imagenetval{}}
            & $\mathbf{\Delta}$
            \\
             \midrule[1.25pt]
            \multirow{2}{*}{MobileNetv3-L} & \datacompTM{}   & \multirow{2}{*}{4.9} &  \multirow{2}{*}{1.1 + 3.3} &  34.1 &\\
             & \ourdatasetTM{} (Ours)& &  & \textbf{44.7} & $\uparrow$\textcolor{Green}{+10.6} \\
            \midrule
            \multirow{2}{*}{ViT-T/16} & \datacompTM{}    & \multirow{2}{*}{5.6} & \multirow{2}{*}{3.0 + 3.3}&  32.9 & \\
             & \ourdatasetTM{} (Ours)  & &  & \textbf{44.1} & $\uparrow$\textcolor{Green}{+11.2} \\
            \midrule
            \multirow{2}{*}{ResNet-50} & \datacompTM{}   & \multirow{2}{*}{24.6} & \multirow{2}{*}{2.6 + 3.3} &  40.4 & \\
             & \ourdatasetTM{} (Ours) & &  & \textbf{51.9} & $\uparrow$\textcolor{Green}{+11.5}\\
            \midrule
            \multirow{2}{*}{FastViT-MA36} & \datacompTM{}   & \multirow{2}{*}{43.5} &  \multirow{2}{*}{4.3 + 3.3} &  45.2 & \\
             & \ourdatasetTM{} (Ours) & & & \textbf{58.9} & $\uparrow$\textcolor{Green}{+13.7} \\
            \bottomrule[1.5pt]
        \end{tabular}
    }
    \caption{\ourdatasetTM{} vs.\ \datacompTM{}. All the models are trained for 30k iterations ($\sim$ 0.24B seen samples).}
    \label{tab:comparison_dr_dataset}
\end{table}

\section{Performance of other models on \ourdatasetTM{}}
In \cref{tab:comparison_dr_dataset}, we compare performance of CLIP models with different sized image encoders when trained on \ourdatasetTM{}. All models enjoy significant accuracy improvement when trained on \ourdatasetTM{} with no training overhead. For example, even the smallest model like MobileNetV3-L with only 4.9M parameters obtains a significant 10.6\% improvement in zero-shot IN-val performance.

\section{Extended Results} \label{sec:extended_results}
In this section we provide extended zero-shot results of our proposed family of CLIP models: \ourmodelSZero{}, \ourmodelSOne{}, \ourmodelSTwo{}, and \ourmodelB{}. 
Zero-shot classification and retrieval results are provided in \cref{tab:full_eval_extended}. 
We also include additional results from related works where only partial evaluation is available.

\begin{table*}[h]
    \centering
    \resizebox{0.99\textwidth}{!}{
        \begin{tabular}{l|ccccccc|cccccc|cccccc}
            \toprule[1.5pt]
            \multirow{3}{*}{\textbf{Name}} 
            & \multicolumn{7}{c}{\textbf{ImageNet Shifts CLS}}
            & \multicolumn{6}{c}{\textbf{\flickrval{} Retrieval}}
            & \multicolumn{6}{c}{\textbf{COCO Retrieval}}
            \\
            \cmidrule(lr){2-8}
            \cmidrule(lr){9-14}
            \cmidrule(lr){15-20}
             &
             \multirow{2}{*}{val}& \multirow{2}{*}{A} & \multirow{2}{*}{R}& \multirow{2}{*}{O}& \multirow{2}{*}{S}& \multirow{2}{*}{V2}& \multirow{2}{*}{Obj}&
             \multicolumn{3}{c}{T$\to$I}&\multicolumn{3}{c}{I$\to$T}&
             \multicolumn{3}{c}{T$\to$I}&\multicolumn{3}{c}{I$\to$T}\\
             \cmidrule(lr){9-11}
             \cmidrule(lr){12-14}
             \cmidrule(lr){15-17}
             \cmidrule(lr){18-20}
             &
             &&&&&&&
             
             @1& @5& @10&
             @1& @5& @10&
             @1& @5& @10&
             @1& @5& @10
             \\
             \midrule[1.25pt]             
             MobileCLIP-B&
             76.8& 58.7& 89.6& 41.4&
             64.5& 69.8& 69.4&
             77.3& 94.4& 96.7&
             91.4& 99.1& 99.9&
             50.6& 74.9& 82.9&
             68.8& 88.3& 92.9
             \\
             MobileCLIP-S2&
             74.4& 49.3& 87.0& 46.9&
             62.2& 66.8& 66.6&
             73.4& 92.3& 95.6&
             90.3& 98.9& 99.6&
             45.4& 70.1& 79.0&
             63.4& 85.1& 91.4
             \\
             MobileCLIP-S1&
             72.6& 40.3 & 84.7& 50.5&
             60.3& 64.9& 63.4&
             71.0 & 91.3& 95.3 &
             89.2 & 98.0 & 99.5 &
             44.0& 68.9& 77.7&
             62.2& 84.3 & 90.1
             \\
             MobileCLIP-S0&
             67.8& 26.5& 78.6& 53.8&
             55.5& 59.9& 55.9&
             67.7& 88.8& 93.3&
             85.9& 97.1& 98.8&
             40.4& 66.0& 75.9&
             58.7& 81.1& 88.2
             \\
             \midrule[1.25pt]
             DIME-FM-B/32~\citep{DIME-FM}&
             66.5& 32.2& 69.8& (-)& 46.5& 58.9& 43.2&
             (-)& (-)& (-)&
             (-)& (-)& (-)&
             (-)& (-)& (-)&
             (-)& (-)& (-)\\
             VeCLIP-B/16~\citep{VeCLIP}&
             64.6& (-)& (-)& (-)& (-)& 57.7& (-)&
             76.3& 93.5& 96.4&
             91.1& 98.5& 99.7&
             48.4& 73.3& 81.8&
             67.2& 87.3& 92.7\\
             TinyCLIP-63M/32~\citep{tinyclip}&
             64.5& 22.8& 74.1& (-)& 50.8& 55.7& 31.2&
             66.0& (-)& (-)&
             84.9& (-)& (-)&
             38.5& (-)& (-)&
             56.9& (-)& (-)\\
             CLIPA-B/16~\citep{CLIPA}&
             63.2& 26.8& 73.2& (-)& 48.7& 55.6& 44.3&
             58.3& (-)& (-)&
             75.3& (-)& (-)&
             35.2& (-)& (-)&
             53.1& (-)& (-)\\
            \bottomrule[1.5pt]
        \end{tabular}
    }
    \caption{Extended zero-shot evaluations. We also include additional results from related works where the full DataComp~\citep{DataComp} evaluation was not accessible. Numbers are read from the corresponding papers. For each method we picked their best model up to ViT-B/16 size. Please see \cref{tab:full_eval} for additional details including runtime benchmarking. Models are sorted by their zero-shot classification performance on \imagenet{}-val. Here our \ourmodelSOne{}
    is fully trained with 13B seen samples.}
    \label{tab:full_eval_extended}
\end{table*}

\section{Long training}\label{sec:long_train}

In \cref{tab:long_train} we provide results for training \ourmodelB{} on more 
than 13B seen samples.  We explore continuing the training of \ourmodelB{} to 
reduce the cost of training from scratch. Recently, \citep{garg2024tic} has 
shown that large scale CLIP models can be continually pretrained as the data 
distribution varies with time. We utilize some of their recipes for continual 
training where we initialize the training with a model previously trained with 
cosine or constant learning rate schedule and restart the training on 
\ourdatasetOneB{}. We utilize a short warmup to stabilize the training and then 
use another constant or cosine learning rate schedule with maximum and minimum 
values equal to the original training. We train using 64 nodes with 8xA100-80GB GPUs 
and a per-GPU batch size of either 128 or 256. One seen sample
for \ourdataset{} is a triplet of one randomly augmented image, one ground-truth caption, and one randomly picked synthetic caption.
Number of iterations is the number of seen samples divided by the global batch size. Note that training wall-clock time is the same for \ourdataset{} vs \datacomp{} (\cref{tab:training_time}).

Compared with our initial training on 13B seen samples, our long training with 
39B total seen samples achieves {0.6\%} improvement in average performance on 
{38} datasets as well as 0.4\% improvement in zero-shot \imagenetval{} accuracy.  
We reach similar improvements in average performance on {38} datasets (0.4\%) 
with only 18B total seen samples by continuing our original training on 13B 
seen samples with a short training using Cosine(40k, 131k, 2k).

\begin{table*}[h!]
    \centering
    \resizebox{0.99\textwidth}{!}{
        \begin{tabular}{l|C{1.5cm}|cccccc|C{1.4cm}}
            \toprule[1.5pt]
            \multirow{2}{2cm}{\centering\textbf{LR Schedule}}
            & \multirow{2}{1.7cm}{\centering\textbf{Seen Samples}}
            & \multicolumn{2}{c}{\textbf{Zero-shot CLS}}
            & \multicolumn{2}{c}{\textbf{\flickrval{} Ret.}}
            & \multicolumn{2}{c}{\textbf{COCO Ret.}}
            & \multirow{2}{1.4cm}{\centering\textbf{Avg. Perf. on 38}}
            \\
            \cmidrule(lr){3-8}
            \cmidrule(lr){5-6}
            \cmidrule(lr){7-8}
            & & \imagenetval{} & IN-shift &T$\to$I&I$\to$T&T$\to$I&I$\to$T& \\
             \midrule[1.25pt]
Cosine(200k, 65k, 2k) & 13B &
76.8 & 65.6 & \textbf{77.3} & 91.4 & \textbf{50.6} & 68.8 & 65.2\\
Const(300k, 65k, 2k) + Cosine(40k, 131k, 2k) & 25B &
\textbf{77.1} & 65.8 & 77.0 & 91.8 & 50.2 & 68.7 & 65.2\\
Const(300k, 65k, 2k) + Cosine(300k, 65k, 2k) & 39B &
\textbf{77.2} & \textbf{66.1} & 76.9 & 92.3 & 50.0 & 68.7 & \textbf{65.8}\\
Const(200k, 65k, 2k) + Cosine(40k, 131k, 2k) & 18B &
\textbf{77.1} & \textbf{65.9} & 77.0 & \textbf{92.8} & 50.3 & \textbf{69.1} & 64.6\\
Cosine(200k, 65k, 2k) + Cosine(40k, 131k, 2k) & 18B &
76.8 & 65.6 & 76.8 & 92.1 & \textbf{50.4} & \textbf{69.1} & \textbf{65.6}\\
Cosine(100k, 131, 2k) + Cosine(40k, 131k, 2k) & 18B &
\textbf{77.0} & 65.6 & \textbf{77.2} & 91.3 & 50.2 & \textbf{69.2} & 64.2\\
            \bottomrule[1.5pt]
        \end{tabular}
    }
    \vspace{-5pt}
    \caption{\textbf{\ourmodelB{} long and continual training.}
    Retrieval performances are reported @1.  Last column shows average 
    performance on 38 datasets as in OpenCLIP~\citep{OpenCLIP}. The learning 
    rate schedules are specified as Cosine/Const(num-iterations, global 
    batch-size, warmup-iterations). We highlight numbers within 0.2\% of the 
    maximum in each column.}
    \label{tab:long_train}
    \vspace{-5pt}
\end{table*}

\end{document}